\documentclass[Afour,sageh,times]{ijrr/sagej}

\usepackage{fontspec}
\usepackage[dvipsnames,table]{xcolor}
\usepackage[colorlinks,bookmarksopen,bookmarksnumbered]{hyperref}
\hypersetup{
    colorlinks=true,
    linkcolor=brown!80!black,
    citecolor=RoyalPurple!90!black,
    urlcolor=Maroon!85!black
}
\setcitestyle{aysep={,}}
\usepackage{amsmath,amssymb,amsthm}
\usepackage[ruled, lined, linesnumbered, commentsnumbered, longend]{algorithm2e}
\usepackage{multirow}
\usepackage{graphicx}
\usepackage{float}
\usepackage{enumitem}
\usepackage{siunitx}
\usepackage{relsize}
\usepackage{xspace}
\usepackage{boldline}
\usepackage{stfloats}
\usepackage{caption}
\usepackage{wasysym}
\usepackage{flushend}
\usepackage{arydshln}
\usepackage{booktabs}
\usepackage{makecell}
\usepackage{tikz}
\usepackage{nicefrac}
\usetikzlibrary{calc,arrows.meta}

\makeatletter
\def\keywords#1{\gdef\@keywords{\begin{minipage}{\textwidth}\sffamily\normalsize\textbf{Keywords}\\#1\end{minipage}}}
\def\abstract{\lrbox\absbox\minipage{\dimexpr\textwidth-9mm\relax}\sagesf\normalsize\section*{\normalsize Abstract}\vskip -1.5mm}
\def\@maketitle{%
  \vspace*{-16pt}\null
  \centering
  \begin{minipage}{\dimexpr\textwidth-9mm\relax}
  {\sffamily\small\itshape Original article\par}\vspace{4pt}\hrule\vspace{11pt}
  \begin{minipage}[t]{\dimexpr\linewidth-49mm\relax}
    {\raggedright\sffamily\titlesize\bfseries\@title\par}
  \end{minipage}\hfill
  \begin{minipage}[t]{45mm}
    \raggedleft\sffamily\scriptsize The International Journal of Robotics Research\\
    \copyright\ The Author(s) \volumeyear\\
    DOI: 10.1177/02783649261511970\\
    journals.sagepub.com/home/ijr
  \end{minipage}
  \par\vspace{27pt}
  {\raggedright\sffamily\large\bfseries\@author\par}
  \vspace{32pt}
  {\noindent\usebox\absbox\par}
  \vspace{19pt}
  {\noindent\@keywords\par}
  \end{minipage}\vspace{26pt}\par}
\def\ps@title{\def\@oddhead{}\let\@evenhead\@oddhead\def\@oddfoot{}\let\@evenfoot\@oddfoot}
\def\ps@sagepage{%
  \let\@mkboth\@gobbletwo
  \def\@evenhead{\parbox{\textwidth}{\normalfont\sffamily\thepage\hfill\itshape\journalname\\[-6pt]\noindent\rule{\textwidth}{0.25pt}}}%
  \def\@oddhead{\parbox{\textwidth}{\normalfont\sffamily\itshape\leftmark\hfill\upshape\thepage\\[-6pt]\noindent\rule{\textwidth}{0.25pt}}}%
  \def\@evenfoot{}\def\@oddfoot{}}
\makeatother
\makeatletter
\def\journalname{The International Journal of Robotics Research}
\makeatother

\newtheorem{theorem}{Theorem}[section]
\newtheorem{proposition}[theorem]{Proposition}

\newtheorem{Assumption}[theorem]{Assumption}
\newtheorem{corollary}[theorem]{Corollary}

\newtheorem{definition}[theorem]{Definition}
\newtheorem{running-example}{Running Example}
\newtheorem*{running-example*}{Running Example}

\newtheorem*{problem*}{Problem}

\makeatletter
\def\endproof{\qed\endtrivlist\@endpefalse}
\makeatother

\newcommand{\longthmtitle}[1]{\mbox{}\textup{\textbf{(#1).}}}
\newcommand{\sbf}[1]{{\small \textbf{#1}}\xspace}

\definecolor{myblue}{HTML}{0014A8}
\definecolor{mygreen}{HTML}{2E6F40}

\newcommand\BibTeX{{\rmfamily B\kern-.05em \textsc{i\kern-.025em b}\kern-.08em
T\kern-.1667em\lower.7ex\hbox{E}\kern-.125emX}}

\def\volumeyear{2026}
\begin{document}

\runninghead{Yu and Liu}

\title{Neural Fidelity Calibration for Informative Sim-to-Real Adaptation}

\author{Youwei Yu\affilnum{1} and
    Lantao Liu\affilnum{1}%
}%
\affiliation{%
    \affilnum{1}Luddy School of Informatics, Computing, and Engineering, Indiana University, Bloomington, IN, USA
}%
\corrauth{Lantao Liu, Luddy School of Informatics, Computing, and Engineering, Indiana University, Bloomington, IN 47408, USA.}
\email{lantao@iu.edu}

\begin{abstract}
Deep reinforcement learning (RL) enables the transfer of agile robotic skills from simulation to the real world. Bridging the sim-to-real gap via domain randomization---training over randomized environments---often requires expert physics knowledge. Prior approaches attempt to close this gap by tuning simulator parameters to match real-world data, but they typically focus on manually selected, low-dimensional subsets. Despite advances, even cutting-edge simulators can miss critical real-world details. Additionally, reconstructing real-world environments in simulation can introduce perception-induced errors that degrade policy improvement.
To address these challenges, we propose Neural Fidelity Calibration (NFC), a framework that uses conditional flow matching to calibrate simulator physical parameters and residual fidelity domains online. Residual fidelity captures sim-to-real dynamic discrepancies and perception-induced environment uncertainty. NFC operates in latent space to efficiently sample realistic environments from inferred posteriors, enabling effective real-to-sim policy fine-tuning.
Our framework is informative and adaptive in three ways: (a) the pretrained policy is fine-tuned online only when an anomaly is detected; (b) sequential NFC seeds its proposal prior from the pretrained NFC, reducing the online training burden of flow matching; and (c) when high NFC uncertainty would hinder policy improvement, a hallucination policy modulates that uncertainty, following the optimism-in-the-face-of-uncertainty principle.
Experimental evaluations demonstrate that our framework achieves superior calibration precision compared to established methods across multiple robots with high-dimensional parameter spaces. Extensive ablation studies reveal the contribution of residual fidelity to policy improvements. Notably, our approach exhibits robust performance in challenging real-world conditions, such as navigating with a broken wheel axle on snowy terrain. The code and demonstration videos are available at \url{https://www.youwei-yu.com/nfc}.
\end{abstract}

\keywords{Sim-to-real adaptation, simulation-based inference, flow matching, field robots}

\maketitle

\section{Introduction}
Zero-shot sim-to-real reinforcement learning enables agile policy transfer across diverse robot embodiments, including soft~\citep{tiboni23rfdroid}, wheeled~\citep{yu2024adaptive}, aerial~\citep{Geles-RSS-24}, and quadruped robots~\citep{Joonho2024wheelleg}. To improve policy robustness under real-world variability, domain randomization (DR) has emerged as a key technique. It bridges the sim-to-real gap by randomizing simulator physical parameters, diversifying environments, and injecting noise or perturbations into the agent's state and observation space~\citep{tobin2017domain, Gleave2020Adversarial}.

Despite these advances, existing methods often rely on expert physics knowledge to select domain randomization ranges~\citep{ma2024dreureka}, reconstruct accurate simulation environments~\citep{escontrela2024learning}, or design realistic perturbation scenarios~\citep{Shi-RSS-24}. Exhaustively sampling simulator parameters and environments is theoretically viable but computationally prohibitive in practice.
An alternative is the sim-to-real-to-sim paradigm: policies are first trained in simulation and then improved with real-world data~\citep{stachowicz2024lifelong}, which also moves risky exploration into simulation. Within this paradigm, simulation-based inference (SBI)~\citep{papamak16efree} narrows the randomization range by inferring simulator parameters from real-world trajectories---for instance, the friction coefficient from a trajectory on snow. Two obstacles remain. First, SBI assumes a well-calibrated simulator, yet even high-fidelity simulators omit real-world effects, and environments reconstructed from onboard sensors inherit perception errors that limit policy improvement. Second, although kernel- and optimization-based methods have shown effectiveness in parameter estimation~\citep{ramos2019bayessim, sobanbabu2025sampling}, they typically operate in low-dimensional parameter spaces and struggle to scale to the high-dimensional problems in reality.

To address these challenges, we propose Neural Fidelity Calibration (NFC), a framework employing conditional flow matching to calibrate simulator physical parameters and residual fidelity domains online during robot operation. Our contributions toward informative sim-to-real policy adaptation include:

\begin{enumerate}[leftmargin=*, label=(\arabic*)]
\item \textbf{Simulator-to-Reality Shift.} We identify two critical concepts contributing to the sim-to-real gap: \emph{simulator calibration} and \emph{residual fidelity}. Simulator calibration optimizes simulator parameters to align robot behavior with real-world dynamics. Residual fidelity captures discrepancies between simulated and real-world dynamics (even when the identified simulator parameters are correct) as well as errors from reconstructed environments due to sensor perception uncertainties.
\item \textbf{Neural Fidelity Calibration.} NFC can efficiently handle high-dimensional calibration by jointly inferring simulator calibration and residual fidelity. The key idea is that NFC employs conditional flow matching operating in a latent space to infer distributions, enabling realistic reconstruction of environments and dynamics.
\item \textbf{Informative Policy Adaptation.} At the policy level, our framework fine-tunes the policy online only when an anomaly indicates the policy is prone to failure. When NFC uncertainty is high, a hallucination policy, jointly optimized with the control policy, modulates that uncertainty following the optimism-in-the-face-of-uncertainty principle. All real-world data is reused offline to reinforce the policy.
\end{enumerate}

We systematically evaluate the effectiveness of NFC against established SBI methods in sim-to-sim and sim-to-real scenarios. Experiments on diverse robots with high-dimensional parameter spaces demonstrate NFC's superior posterior inference, and both simulated and real-world experiments confirm that the residual fidelity plays a pivotal role in policy performance under challenging conditions.

\section{Related Work}\label{sec:related}
In the realm of sim-to-real agile skill transfer, we first review progress in robust behavior learning and focus on adaptive domain randomization, which calibrates the simulator to mirror the real-world environment. Then we discuss how to fine-tune the RL policy under uncertain calibration.

\subsection{Robust Sim-to-Real Transfer}
Zero-shot sim-to-real transfer with reinforcement learning has proven effective in agile locomotion~\citep{David24anymalparkour, zhuang2024humanoid}, navigation~\citep{Fabian24DTC, yu2024adaptive}, and dexterous manipulation~\citep{Handa23DeXtreme, Villasevil-RSS-24}, where domain randomization is essential for bridging the sim-to-real gap. However, relying on repeated domain randomization alone can be inefficient and degrade robustness, prompting researchers to explore adversarial attacks and safe reinforcement learning.

Adversarial agents often exploit privileged information to reveal policy vulnerabilities. They can directly apply destabilizing forces and alter physical parameters in simulation, or mimic normal agents while aiming to induce failures~\citep{Gleave2020Adversarial}. They can also be virtual adversaries that feed deceptive states to the agent, with vision-based inputs being especially susceptible~\citep{lin2017tactics}. \cite{Pattanaik2018Attacks} sampled worst-performing trajectories by tricking the agent into poor decisions, while \cite{NEURIPS2021_dbb42293} considered adversarial attacks under worst-case reward sequences. Moreover, \cite{Shi-RSS-24} used sequential attacks to show that domain randomization alone is insufficient against adversaries but can be fortified by adversarial training.

Safe reinforcement learning seeks to balance exploratory behavior with safety requirements. To enforce constraints, constrained Markov decision processes (MDPs) incorporate risk measures such as conditional value at risk~\citep{Stachowicz-RSS-24}, and Lagrangian methods transform constrained objectives into unconstrained ones~\citep{JMLR:v18:15-636}. Other approaches encode safety via Hamilton-Jacobi-Bellman-Isaacs variational inequalities~\citep{Hsu-RSS-21} and control barrier functions~\citep{Cheng2019safe}. Sim-to-Lab-to-Real~\citep{hsuzen2022sim2lab2real} used Hamilton-Jacobi reachability to train a backup (safety) policy and a probably approximately correct (PAC) Bayes framework to guarantee performance and safety. More recently, \cite{nguyen2024gameplay} proposed a predictive safety filter that simulates adversarial scenarios to preempt failures and ensure safe exploration.
However, the guarantees of these methods hold relative to their own assumptions---the assumed model, the admissible disturbance set, the available state information, and constraint feasibility---any of which a structural change such as an unmodeled fault may violate. This motivates continual real-world recalibration, triggered when an anomaly is detected.

\subsection{Adaptive Domain Randomization}
Adaptive domain randomization (ADR) aligns simulator parameters with real-world robot executions, providing insights into physical phenomena while narrowing the scope of domain randomization. Thanks to its theoretical generality, ADR has been applied to material cutting~\citep{Heiden-RSS-21}, soft robot control~\citep{tiboni23rfdroid}, and robot manipulation~\citep{huang2023went, ramos2019bayessim}.

Most existing methods focus on improving the likelihood-free simulation-based inference (SBI) framework~\citep{papamak16efree}. For instance, \cite{ramos2019bayessim} used random Fourier features to highlight the sensitivity of approximate Bayesian computation (ABC) to distance metrics, and \cite{pmlr-v164-muratore22a} introduced sequential inference for real-world scenarios, though the potential for high-dimensional time-series or image data remains under-explored~\citep{greenberg2019automatic}. \cite{huang2023went} proposed differentiable causal factors to expose the sim-to-real gap. From an information-theoretic perspective, \cite{memmel2024asid} maximized Fisher information to identify trajectories highly sensitive to unknown parameters. Further, \cite{josif2024continual} integrated continual learning but only sampled randomization parameters in small subsets. 

Beyond inference-based methods, optimization techniques like evolution strategies~\citep{tsai21droid, tiboni23rfdroid} and Bayesian optimization~\citep{ota21augsim} have also shown promise, albeit with higher computational costs. \cite{pmlr-v229-ren23b} highlighted the value of meta-learning~\citep{wang2016learning} for further adaptation. Despite their general applicability, these methods still suffer from limited simulation budgets and mismatched simulation priors relative to real-world conditions.

Alternatively, optimization-based frameworks can bypass some of these constraints but often require differentiable simulators. For instance, a differentiable simulator for soft material cutting~\citep{Heiden-RSS-21} achieved sharper inference via stochastic gradient Langevin dynamics (SGLD) compared to a non-differentiable approach~\citep{ramos2019bayessim}, and improvements were demonstrated with Stein variational gradient descent (SVGD) in manipulator mass inference \citep{Heiden21neuralsim, Heiden22inf}. Two limitations remain: these approaches often require extensive iterative simulations to achieve convergence, and the idea of a perfectly calibrated simulator that precisely mirrors real-world dynamics remains elusive.

\subsection{Optimistic Exploration under Uncertainty}
A theoretically grounded approach to guide exploration, which has been extensively studied in single-agent RL, is the well-known \emph{optimism in the face of uncertainty} (OFU) principle. In short, it chooses actions that maximize an optimistic estimate of the agent's value function. OFU can efficiently balance exploration with exploitation and has been applied in several single-agent RL domains~\citep{brafman2002rmax, JMLR:v11:jaksch10a}.

Hallucinated upper confidence reinforcement learning (H-UCRL)~\citep{curi2020efficient} modified the dynamics model with hallucinated controls to find the policy with the greatest cumulative rewards under epistemic uncertainty. H-MARL~\citep{sessa2022efficient} extended it to the multi-agent optimistic hallucinated game that seeks the coarse correlated equilibrium. Separately, \cite{sukhija2023optimistic} incorporated the information gain between the unknown dynamics and state observations into the optimal exploration objectives. With the dynamics model bootstrapped from model-based meta RL, H-UCRL could guide exploration in a data-efficient fashion~\citep{bhardwaj24metarl}. The regularization and epistemic uncertainty quantification were incorporated in both the meta-learning and task adaptation stages. Instead of pessimistic (safe) constraints that support only a subset of all possible valid plans, the optimistic symbolic planner~\citep{sreedharan2023optimistic} allows for more plans than are possible in the true model.
Our hallucination policy (Sec.~\ref{sec:hallucinate}) borrows this optimism but uses a different mechanism: rather than optimizing within a learned model's confidence set, it reshapes the spread of simulated domains drawn from the NFC posterior, and therefore inherits no confidence-bound guarantee.

\section{Preliminaries}\label{sec:prelim}
We review simulation-based inference (SBI), a framework widely employed for parameter estimation from real-world data using computationally inexpensive simulations. We then introduce flow matching as a state-of-the-art posterior estimator, particularly suitable for high-dimensional spaces. Building upon these foundations, we formulate the sim-to-real policy transfer problem, emphasizing scenarios where policies initialized in simulation require improvement through real-world deployment.

\subsection{Simulation-Based Inference}
Although domain randomization (DR) can mitigate the sim-to-real gap, it typically demands expert knowledge of real-world physics~\citep{ma2024dreureka} to achieve robust behavior. Adaptive domain randomization~\citep{ramos2019bayessim}, by contrast, infers simulator parameters $ \boldsymbol{\psi} $ (e.g., friction, mass) from real-world experiences, iteratively ``calibrating'' the simulator through Bayesian inference $ p(\boldsymbol{\psi}\vert \boldsymbol{\tau}) \propto p(\boldsymbol{\tau}\vert \boldsymbol{\psi}) p(\boldsymbol{\psi}) $, where $ \boldsymbol{\tau} $ represents a real-world trajectory (e.g., state--action history). This approach belongs to the family of simulation-based inference (SBI). SBI samples parameters from a prior domain distribution $ p(\boldsymbol{\psi}) $, simulates the corresponding trajectories, compares them against real executions, and infers the posterior distribution $ p(\boldsymbol{\psi}\vert \boldsymbol{\tau}) $. By narrowing the DR space, SBI enhances real-to-sim policy fine-tuning efficiency.

We assume access to a parallelizable, faster-than-real-time black-box simulator capable of modeling complex robotic dynamics. The key challenge lies in computing the likelihood $ p(\boldsymbol{\tau}\vert \boldsymbol{\psi}) $, which usually requires a differentiable simulator. Likelihood-free inference methods in SBI, notably approximate Bayesian computation (ABC), address this by approximating the posterior distribution without explicit likelihood calculations. ABC infers parameters by comparing simulated data with real observations, typically based on observation discrepancy measures. Common ABC variants include:
\begin{enumerate}[leftmargin=*, label=(\arabic*)]
\item Rejection ABC approximates the posterior via $ p(\|\boldsymbol{\tau}^{\mathrm{sim}} - \boldsymbol{\tau}^{\mathrm{real}}\| < \epsilon \vert \boldsymbol{\psi}) $ with tolerance $ \epsilon \geq 0 $ using Monte Carlo simulations.
\item Sequential Monte Carlo ABC~\citep{Sisson2007SMC} dynamically updates the choice of $ \epsilon $, which balances the estimation accuracy and computation time.
\item $ \epsilon $-free ABC~\citep{papamak16efree} approximates the posterior $ \hat{p}(\boldsymbol{\psi}\vert \boldsymbol{\tau}) \propto \frac{ p(\boldsymbol{\psi}) }{ \tilde{p}(\boldsymbol{\psi}) } q(\boldsymbol{\psi}\vert \boldsymbol{\tau}) $ without a tolerance constraint. It estimates the conditional density $ q(\boldsymbol{\psi}\vert \boldsymbol{\tau}) $, where the proposal prior $ \tilde{p}(\boldsymbol{\psi}) $ describes the initial distribution of $ \boldsymbol{\psi} $.
\end{enumerate}
These ABC methods iteratively refine posterior estimates until convergence, usually based on some trajectory discrepancy measures. Despite their effectiveness, they become computationally intensive as simulator parameters $ \boldsymbol{\psi} $ grow in dimensionality or complexity, requiring extensive repeated simulations. Meanwhile, neural-network and kernel-based estimators typically target low-dimensional settings, e.g., a handful of physical parameters~\citep{ramos2019bayessim} or the mass--inertia matrix~\citep{sobanbabu2025sampling}. 

The problems considered in our work instead couple up to \num{39} physical parameters with residual-fidelity variables whose environment component alone is $\num{16384}$-dimensional, conditioned on trajectories that include depth or elevation perception (Sec.~\ref{sec:sim-exp}).

\subsection{Neural Posterior Inference}\label{sec:neural-posterior}
Uniform priors are commonly used to initiate ABC inference; however, they typically require extensive simulations, limiting real-world online adaptation. Moreover, robotics scenarios characterized by rich observational data (e.g., depth images) pose challenges to standard neural-network-based inference methods~\citep{pmlr-v202-geffner23a}, particularly when the simulator parameter $ \boldsymbol{\psi} $ is also high-dimensional. We therefore estimate the posterior with a conditional continuous-time generative model, instantiated as conditional flow matching~\citep{lipman2022flow}, which scales to high-dimensional targets and rich conditioning~\citep{wildberger2023flow}.

Flow matching~\citep{lipman2022flow} learns a time-dependent velocity field that transports a simple base distribution to the target. Let $ \boldsymbol{\psi}(0) \sim \mathcal{N}(0, I) $ and let $ \boldsymbol{\psi}(1) $ be a parameter sample. Along the conditional probability path $ \boldsymbol{\psi}(k) = (1-k)\, \boldsymbol{\psi}(0) + k\, \boldsymbol{\psi}(1) $, $ k \in [0, 1] $, the target velocity is $ \boldsymbol{\psi}(1) - \boldsymbol{\psi}(0) $. The conditional flow-matching objective over the network parameters $ \boldsymbol{\theta} $ of the velocity field $ \boldsymbol{v}_{\boldsymbol{\theta}} $ is
\begin{equation}\label{eqn:flow-matching}
\begin{aligned}
\boldsymbol{\theta}^{*} = & \operatorname*{argmin}_{\boldsymbol{\theta}} \mathbb{E}_{k \sim \mathcal{U}(0,1)} \mathbb{E}_{\boldsymbol{\psi}(0), \boldsymbol{\psi}(1)} \big[ \\
& \left\| \boldsymbol{v}_{\boldsymbol{\theta}}(\boldsymbol{\psi}(k), \boldsymbol{\tau}, k) - \left( \boldsymbol{\psi}(1) - \boldsymbol{\psi}(0) \right) \right\|_{2}^{2} \big],
\end{aligned}
\end{equation}
where the trajectory $ \boldsymbol{\tau} $ conditions the velocity field.

To draw samples from $ p(\boldsymbol{\psi}\vert \boldsymbol{\tau}^{\mathrm{real}}) $, we integrate the learned velocity field from $ k = 0 $ to $ k = 1 $,
\begin{equation}\label{eqn:flow-ode}
\frac{\mathrm{d} \boldsymbol{\psi}(k)}{\mathrm{d} k} = \boldsymbol{v}_{\boldsymbol{\theta}}(\boldsymbol{\psi}(k), \boldsymbol{\tau}^{\mathrm{real}}, k), \quad \boldsymbol{\psi}(0) \sim \mathcal{N}(0, I),
\end{equation}
starting from a Gaussian sample. Continuous-time generative estimators of this kind capture posterior distributions more accurately than standard neural networks or kernel methods, particularly with limited data~\citep{sharrock2024sequential, wildberger2023flow}; flow matching additionally admits a deterministic ODE sampler, so a fixed, small number of integration steps suffices for online use. Throughout, we write $ \boldsymbol{q}_{\boldsymbol{\theta}} $ for the conditional density induced by Eq.~\eqref{eqn:flow-ode}.

\begin{figure*}[t]
\centering
{\setlength{\fboxsep}{0pt}\setlength{\fboxrule}{0.4pt}%
\fbox{\resizebox{\dimexpr\linewidth-2\fboxrule\relax}{!}{%
  \input{media/fig_sbi_nfc.tex}%
}}}
\caption{
Simulation-based inference (SBI) instantiated by Neural Fidelity Calibration. A simulator and a uniform prior generate paired training examples for a conditional flow matching estimator. The observed rollout $\boldsymbol{\tau}^{\mathrm{real}}$ is encoded and conditions the transport from the base distribution at $k=0$ to the calibrated joint posterior at $k=1$, from which the residual fidelity $g_{\boldsymbol{\phi}}(\cdot)$ explains the sim-to-real discrepancy. The red cross denotes the ground-truth parameter.
}
\label{fig:sbi-nfc-overview}
\end{figure*}

Figure~\ref{fig:sbi-nfc-overview} instantiates this inference pipeline: a simulator and a uniform prior generate paired examples, and the encoded rollout conditions the transport from the base distribution to the posterior. The residual-fidelity component shown on the right is introduced in our method.

\subsection{Problem Formulation}
We model our framework as a contextual Markov decision process~\citep{hallak2015cmdp}, defined by the tuple $(\mathcal{S}, \mathcal{A}, r, \gamma, p, \mathcal{C})$, where $\mathcal{S}$ is the state space, $\mathcal{A}$ is the action space, $\gamma \in (0, 1]$ is the discount factor, $r : \mathcal{S} \times \mathcal{A} \rightarrow \mathbb{R}$ is the reward function, and $p_{c}(s_{t+1}| s_t, a_t)$ denotes the dynamics for state $s \in \mathcal{S}$, action $a \in \mathcal{A}$, and context $c \in \mathcal{C}$. The contextual space $\mathcal{C}$ encompasses both simulator parameters $\boldsymbol{\psi}$ (e.g., friction, mass) and environment variables $\boldsymbol{e}$ (e.g., terrain geometry). Our objective is to optimize the policy $\boldsymbol{\pi}: \mathcal{S} \rightarrow \mathcal{A}$ under a context distribution inferred from real-world execution:
\begin{equation}\label{eqn:problem-form}
    \max_{\boldsymbol{\pi}} \; \mathbb{E}_{c \sim p(c \vert \boldsymbol{\tau}^{\mathrm{real}})} \, \mathbb{E}_{\boldsymbol{\tau} \sim P^{\boldsymbol{\pi}}_{c}} \left[ R(\boldsymbol{\tau}) \right],
\end{equation}
where $P^{\boldsymbol{\pi}}_{c}$ is the distribution of trajectories obtained by rolling out $\boldsymbol{\pi}$ under dynamics $p_{c}$, $R(\boldsymbol{\tau}) = \sum_{t=0}^{T} \gamma^t r(s_{t}, a_{t})$ is the cumulative reward, and $p(c \vert \boldsymbol{\tau}^{\mathrm{real}})$ is the posterior over contexts given the executed real-world trajectory.

The expectation over $p(c \vert \boldsymbol{\tau}^{\mathrm{real}})$ in Eq.~\eqref{eqn:problem-form} is deliberate. The alternative---calibrating the simulator to a single setting $\boldsymbol{\psi}^{*}$ that minimizes a trajectory discrepancy $\mathcal{L}(\boldsymbol{\tau}^{\mathrm{real}}, \boldsymbol{\tau}^{\mathrm{sim}})$---is prone to inaccurate estimation or overfitting for two reasons: (1) simulators cannot replicate every detail of the real world, so no single $\boldsymbol{\psi}$ explains the data; and (2) real-world data is scarce and costly to collect, so the discrepancy is evaluated on few trajectories. A policy fine-tuned in such a simulator can produce unsafe behavior in the real world. Averaging over a posterior instead trains the policy on the range of contexts the model finds consistent with the real observation.

Since policies can be readily initialized in simulation, we target informative fine-tuning during real-world deployment rather than exhaustive pretraining. The reason is practical: simulation budgets are finite. In our prior work~\citep{yu2024adaptive}, a diffusion model generated millions of terrains for training, but under a fixed budget each terrain received only a few draws of the physical parameters, so the joint terrain--dynamics space was sparsely covered. Inferring the posterior $p(c \vert \boldsymbol{\tau}^{\mathrm{real}})$ online, and fine-tuning against it, is the subject of our framework.

\section{Method}\label{sec:method}
\begin{figure*}[t!]
\vspace*{0.04in}
\centering
{\setlength{\fboxsep}{0pt}\setlength{\fboxrule}{0.4pt}\fbox{\includegraphics[width=\dimexpr\linewidth-2\fboxrule\relax]{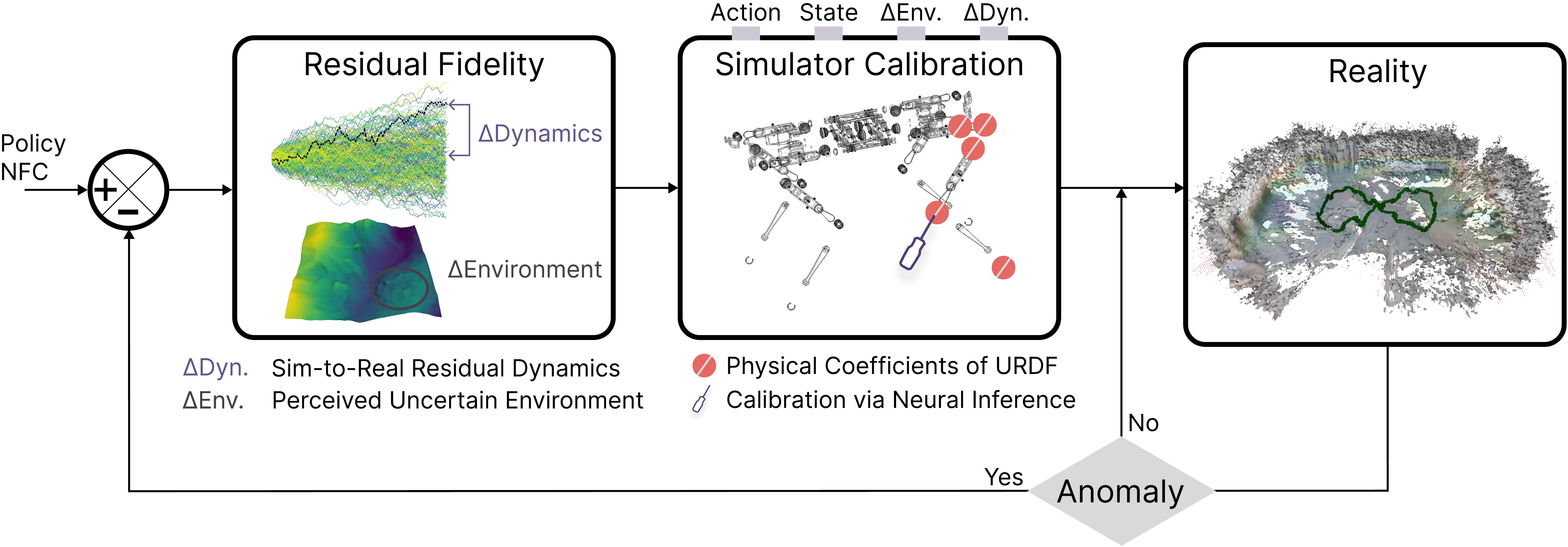}}}
\caption{
Overview of NFC-driven sim-to-real adaptation. (a)~\textbf{Residual fidelity}: colored trajectories are rollouts in the calibrated simulator; black is the real-world execution; their difference is the residual dynamics. The circled region on the digital terrain map marks uncertain perception, where NFC samples the residual environment (ground-truth minus perceived elevation) and reconstructs multiple terrain variants for fine-tuning. (b)~\textbf{Simulator calibration}: a flow matching model infers physical parameters that match the real trajectory. (c)~\textbf{Informative adaptation}: the policy and NFC are pretrained in simulation and fine-tuned only when an anomaly is detected.
}
\label{fig:system}
\end{figure*}

In this section, we present Neural Fidelity Calibration (NFC), designed to bridge the sim-to-real gap effectively. We first provide a general formulation of NFC, highlighting the sim-to-real gaps of simulator calibration and residual fidelity. Second, we introduce sequential NFC, which iteratively updates the posterior distributions online, and latent NFC, which addresses computational challenges associated with high-dimensional data by operating in a reduced latent space. Third, because high NFC uncertainty can stall policy fine-tuning, we integrate an optimistic exploration strategy that lets the agent ``hallucinate'' a modulated amount of randomness. Last, as depicted in Fig.~\ref{fig:system}, we embed NFC in a deployment protocol for online learning.

\subsection{Neural Fidelity Calibration}
Simulation-based inference compares synthetic trajectories generated under sampled parameters with the real execution. BayesSim~\citep{ramos2019bayessim} and its successors rely on Bayesian inference of the form
\begin{equation}
    p(\boldsymbol{\psi}\vert \boldsymbol{\tau}^{\mathrm{real}}) \propto p(\boldsymbol{\tau}^{\mathrm{real}}\vert \boldsymbol{\psi}) p(\boldsymbol{\psi}),
\end{equation}
which can only explain the real trajectory through the simulator's own parameterization, i.e., it assumes $p(\boldsymbol{\psi} \vert \boldsymbol{\tau}^{\mathrm{real}}) \approx p(\boldsymbol{\psi} \vert \boldsymbol{\tau}^{\mathrm{sim}})$ for some $\boldsymbol{\psi}$. This assumption fails in two ways. Even high-fidelity simulators omit real-world effects, so no value of $\boldsymbol{\psi}$ closes the gap; and environments reconstructed from onboard sensors inherit perception errors, so the simulator can be calibrated against the wrong terrain. We therefore separate two sources of discrepancy: calibration shifts and fidelity shifts.

\begin{definition}\longthmtitle{Calibration and Fidelity Shifts}
We define two distinct sources of discrepancy between simulation and reality. (1) Calibration shifts signify that specific parameters are inadequately calibrated for simulating real-world scenarios. (2) Fidelity shifts imply that certain modeling factors are overlooked or missed in the abstract simulator. 

Formally, the real-world transition dynamics can be decomposed as:
\begin{equation}\label{eqn:sys-dynamics}
    s_{t+1} = \color{myblue}{\underbrace{\color{black}{f_{\color{myblue}{\boldsymbol{\psi}}}(s_t, a_t\vert \boldsymbol{e})}}_{\mathrm{Calibration}} \;{\color{black}{+}}\; \underbrace{\color{black}{g_{\color{myblue}{\boldsymbol{\phi}}}(s_t, a_t\vert \boldsymbol{e})}}_{\mathrm{Fidelity}} \color{black}{\;+\; \epsilon(s_t, a_t\vert \boldsymbol{e})}} \color{black}, 
\end{equation}
where $f_{\boldsymbol{\psi}}$ is the nominal simulator dynamics parameterized by $\boldsymbol{\psi}$ (e.g., mass, friction), $g_{\boldsymbol{\phi}}$ models residual effects due to unmodeled dynamics and perception errors, and $\epsilon$ is the noise.
\end{definition}

In robotics, calibration shifts (via $\boldsymbol{\psi}$) directly influence how physical parameters affect robot motion. For example, different friction parameters produce distinct state evolutions from the same initial conditions. Fidelity shifts (via $\boldsymbol{\phi} = \{\Delta s_{t+1}, \Delta \boldsymbol{e}\}$) explain residual discrepancies not accounted for by calibration alone. Specifically, residual dynamics ($\Delta s_{t+1}$) reflect state differences between simulation and reality even under identical conditions (i.e., when the simulator cannot capture all real-world factors), while residual environment ($\Delta \boldsymbol{e}$) captures differences between perceived and true environments due to sensor noise and perception errors. To formulate our framework from a Bayesian inference perspective, we begin with the following assumption.

\begin{Assumption}\label{asm:support}
Since every physical parameter has a bounded plausible range, we assume throughout that $ \boldsymbol{\psi} \in \boldsymbol{\Psi} \subseteq \mathbb{R}^{|\boldsymbol{\Psi}|} $, $ \boldsymbol{\phi} \in \boldsymbol{\Phi} \subseteq \mathbb{R}^{|\boldsymbol{\Phi}|} $, and $ \boldsymbol{\tau} \in \boldsymbol{\Gamma} \subseteq \mathbb{R}^{|\boldsymbol{\Gamma}|} $ are random variables supported on compact domains, each equipped with bounded probability density functions:
\begin{equation}
\begin{aligned}
    p(\boldsymbol{\psi}) > 0,\ \boldsymbol{\psi} \in \boldsymbol{\Psi} \quad &\text{and} \quad p(\boldsymbol{\psi}) = 0,\ \boldsymbol{\psi} \notin \boldsymbol{\Psi}, \\
    p(\boldsymbol{\phi}) > 0,\ \boldsymbol{\phi} \in \boldsymbol{\Phi} \quad &\text{and} \quad p(\boldsymbol{\phi}) = 0,\ \boldsymbol{\phi} \notin \boldsymbol{\Phi}, \\
    p(\boldsymbol{\tau}) > 0,\ \boldsymbol{\tau} \in \boldsymbol{\Gamma} \quad &\text{and} \quad p(\boldsymbol{\tau}) = 0,\ \boldsymbol{\tau} \notin \boldsymbol{\Gamma}.
\end{aligned}
\end{equation}
\end{Assumption}

This bounded support prevents learning implausible simulator parameters or fidelity shifts. For instance, gravity is constrained to a realistic range, and physical damage is restricted to valid configurations (e.g., no more than two broken wheels on a four-wheeled robot, as full immobility would render the policy ineffective).
NFC requires broad, physically admissible support bounds together with a choice of how perturbations enter the dynamics and observations. It does not require real-world anomaly labels or a taxonomy of faults.

Given this assumption, we jointly infer the simulator calibration parameters $\boldsymbol{\psi}$ and residual fidelity parameters $\boldsymbol{\phi}$ using Bayesian inference. The posterior is defined as:
\begin{equation}\label{eqn:neural-posterior}
    p(\boldsymbol{\psi}, \boldsymbol{\phi}\vert \boldsymbol{\tau}) \propto p(\boldsymbol{\tau}\vert \boldsymbol{\psi}, \boldsymbol{\phi}) p(\boldsymbol{\psi}, \boldsymbol{\phi}),
\end{equation}
where $ \boldsymbol{\tau} = \{(s_t, a_t)\}_{t=0}^{H} $ denotes the trajectory collected by executing the policy $ \boldsymbol{\pi}(\cdot) $ over a history of $ H $ steps. Directly computing or maximizing the likelihood $p(\boldsymbol{\tau} \vert \boldsymbol{\psi}, \boldsymbol{\phi})$ is generally intractable due to the ``black-box'' simulator. 
A natural surrogate is a trajectory discrepancy $ \mathcal{L}(\boldsymbol{\tau}^{\mathrm{sim}}, \boldsymbol{\tau}^{\mathrm{real}}) $, such as the squared error between simulated and real states, minimized over $(\boldsymbol{\psi}, \boldsymbol{\phi})$.

We, however, take a different route and learn the posterior itself: a conditional density estimator is trained on parameter--trajectory pairs generated by the simulator, maximizing the average log-density:
\begin{equation}\label{eqn:fidelity-loss-func}
    \boldsymbol{\theta}^* = \operatorname*{argmax}_{\boldsymbol{\theta}} \frac{1}{N} \sum_{n} \log \boldsymbol{q}_{\boldsymbol{\theta}}(\boldsymbol{\psi}_{n}, \boldsymbol{\phi}_{n}\vert \boldsymbol{\tau}_n),
\end{equation}
which recovers the posterior over parameters that explain the observed trajectories (Proposition~\ref{prop:proposal}); the trajectory discrepancy enters implicitly through the simulated data-generating process. Fig.~\ref{fig:sbi-nfc-overview} summarizes the resulting pipeline: the simulator and the prior of Assumption~\ref{asm:support} generate paired examples, the encoded rollout conditions the flow from the base distribution to the joint posterior over $\boldsymbol{\psi}$ and $\boldsymbol{\phi}$, and the residual fidelity $g_{\boldsymbol{\phi}}$ accounts for what the calibrated simulator leaves unexplained.
We implement $\boldsymbol{q}(\cdot)$ with conditional flow matching (Sec.~\ref{sec:neural-posterior}), training a velocity field
\begin{equation}\label{eqn:nfc-velocity}
\boldsymbol{v}_{\boldsymbol{\theta}}\big( [\boldsymbol{\psi}(k), \boldsymbol{\phi}(k)],\, \boldsymbol{\tau}^{\mathrm{real}},\, k \big),
\end{equation}
where $k \in [0,1]$ is the flow time. Integrating Eq.~\eqref{eqn:flow-ode} from a base sample yields an endpoint whose conditional density we denote $\boldsymbol{q}_{\boldsymbol{\theta}}(\boldsymbol{\psi},\boldsymbol{\phi} \vert \boldsymbol{\tau}^{\mathrm{real}})$.

\begin{running-example}
\emph{
We have introduced the abstract concept of Neural Fidelity Calibration, which accounts for both simulator calibration and residual fidelity shifts. To clarify its implementation, we illustrate it in the context of a wheeled robot navigating unstructured terrain. Let the perceived environment be $ \hat{\boldsymbol{e}} \in \mathbb{R}^{R \times C} $, an elevation map obtained from depth sensors. The key parameter sets are as follows:
\begin{enumerate}[leftmargin=*, label=(\arabic*)]
\item Simulator Calibration: $ \boldsymbol{\psi} \in \mathbb{R}^{13} $ represents physical parameters, including the masses of the main body and four wheels $ (\mathbb{R}^{5}) $, motor damping of four wheel joints $ (\mathbb{R}^{4}) $, and friction of four wheels $ (\mathbb{R}^{4}) $.
\item Residual Fidelity: $ \boldsymbol{\phi} = \{\Delta \boldsymbol{s}, \Delta \boldsymbol{e}\} \in \mathbb{R}^{13 + R \times C} $ captures the residual dynamics $ \Delta s_{t+1} = s_{t+1} - f_{\boldsymbol{\psi}}(s_t, a_t) \in \mathbb{R}^{13} $ and the residual environment $ \Delta \boldsymbol{e} = \boldsymbol{e} - \hat{\boldsymbol{e}} \in \mathbb{R}^{R \times C} $. The residual state comprises linear velocity $ (\mathbb{R}^{3}) $, angular velocity $ (\mathbb{R}^{3}) $, projected gravity $ (\mathbb{R}^{3}) $, goal position in the robot frame $ (\mathbb{R}^{3}) $, and heading $ (\mathbb{R}^{1}) $. $ \hat{\boldsymbol{e}} $ and $ \boldsymbol{e} $ represent the perceived and reconstructed terrain elevations, respectively.
\end{enumerate}
During online adaptation, NFC conditions on the robot trajectory $ \boldsymbol{\tau} $ and perceived elevation map $ \hat{\boldsymbol{e}} $ to infer posterior samples of $ \boldsymbol{\psi} $ and $ \boldsymbol{\phi} $. For real-to-sim fine-tuning, we sample $ N $ parameter pairs $ \{\boldsymbol{\psi}_n, \boldsymbol{\phi}_n \}_{n=1}^{N} $ and reconstruct corresponding simulated environments $ \boldsymbol{e} = \hat{\boldsymbol{e}} + \Delta \boldsymbol{e} $. We use them to reset the simulator across training episodes, since $ N $ usually equals the number of parallel simulations. This enables policy refinement under environments that closely reflect the inferred real-world conditions.
}
\end{running-example}

\subsection{Sequential Neural Fidelity Calibration}
Building on the NFC framework, we extend it to a sequential setting for online adaptation. In real-world deployment, policy adaptation requires continuous learning while balancing performance improvement and computational efficiency. Executing a policy without fast model updates poses significant risk. To enable fast and data-efficient adaptation, we formulate an online variant of NFC that incrementally refines its posterior distribution over simulator and fidelity parameters after each trajectory.

At the start of deployment, we leverage the flow matching estimator pretrained in simulation that serves as the initial proposal prior:
\begin{equation}
\tilde{p}(\boldsymbol{\psi}, \boldsymbol{\phi}) = \boldsymbol{q}^{\mathrm{sim}}(\boldsymbol{\psi}, \boldsymbol{\phi} \vert \boldsymbol{\tau}^{\mathrm{real}}),
\end{equation}
where $ \boldsymbol{q}^{\mathrm{sim}} $ is the flow matching model pretrained in simulation, i.e., the simulation-pretrained posterior conditioned on the current real trajectory serves as the proposal for the next round. This initialization captures meaningful parameters and allows faster convergence when encountering real-world trajectories. Compared with a broad prior such as that in Assumption~\ref{asm:support}, it concentrates online simulation on plausible parameters and therefore shortens each adaptation cycle. Using a proposal prior raises one question: a density estimator trained on proposal-generated data no longer targets the true posterior $p(\boldsymbol{\psi}, \boldsymbol{\phi} \vert \boldsymbol{\tau})$ but a reweighted version of it. The following proposition, inspired by \cite{papamak16efree}, makes the relationship explicit; the corollary recovers the standard case.

\begin{proposition}\label{prop:proposal}
Let $ \{(\boldsymbol{\psi}_n, \boldsymbol{\phi}_n, \boldsymbol{\tau}_n)\}_{n=1}^{N} $ be independent and identically distributed samples, each generated by drawing parameters from a proposal prior $ \tilde{p}(\boldsymbol{\psi}, \boldsymbol{\phi}) $ and then a trajectory from the likelihood $ p(\boldsymbol{\tau} \vert \boldsymbol{\psi}, \boldsymbol{\phi}) $:
\begin{equation}\label{eqn:gen-iid}
(\boldsymbol{\psi}_n, \boldsymbol{\phi}_n) \sim \tilde{p}(\boldsymbol{\psi}, \boldsymbol{\phi}) \quad \text{and} \quad \boldsymbol{\tau}_n \sim p(\boldsymbol{\tau} \vert \boldsymbol{\psi}_n, \boldsymbol{\phi}_n).
\end{equation}
Then, the learned posterior probability $ \prod_n \boldsymbol{q}(\boldsymbol{\psi}_n, \boldsymbol{\phi}_n \vert \boldsymbol{\tau}_n) $ is maximized over all conditional densities $ \boldsymbol{q} $ (as $ N \to \infty $) if and only if:
\begin{equation}
\boldsymbol{q}(\boldsymbol{\psi}, \boldsymbol{\phi} \vert \boldsymbol{\tau}) \propto \frac{\tilde{p}(\boldsymbol{\psi}, \boldsymbol{\phi})}{p(\boldsymbol{\psi}, \boldsymbol{\phi})} p(\boldsymbol{\psi}, \boldsymbol{\phi} \vert \boldsymbol{\tau}).
\end{equation}
\end{proposition}

\begin{proof}
By Eq.~\eqref{eqn:gen-iid}, each sample follows the joint density $ (\boldsymbol{\psi}_n, \boldsymbol{\phi}_n, \boldsymbol{\tau}_n) \sim \tilde{p}(\boldsymbol{\psi}, \boldsymbol{\phi}) p(\boldsymbol{\tau} \vert \boldsymbol{\psi}, \boldsymbol{\phi}) $. Due to the strong law of large numbers, as $ N \to \infty $, the average log probability converges almost surely to the following expectation:
\begin{equation}
\begin{aligned}
&\frac{1}{N} \sum_{n} \log \boldsymbol{q}(\boldsymbol{\psi}_{n}, \boldsymbol{\phi}_{n}\vert \boldsymbol{\tau}_n) \\\xrightarrow{a.s.}& \mathbb{E}_{\tilde{p}(\boldsymbol{\psi}, \boldsymbol{\phi}) p(\boldsymbol{\tau} \vert \boldsymbol{\psi}, \boldsymbol{\phi})} \left[ \log \boldsymbol{q}(\boldsymbol{\psi}, \boldsymbol{\phi} \vert \boldsymbol{\tau}) \right].
\end{aligned}
\end{equation}

Let $ \tilde{p}(\boldsymbol{\tau}) = \iint \tilde{p}(\boldsymbol{\psi}, \boldsymbol{\phi}) p(\boldsymbol{\tau} \vert \boldsymbol{\psi}, \boldsymbol{\phi}) \mathrm{d}\boldsymbol{\psi} \mathrm{d}\boldsymbol{\phi} $ denote the trajectory marginal under the proposal. Maximizing the above expectation w.r.t. $ \boldsymbol{q} $ is equivalent to minimizing the following KL divergence:
\begin{equation}
\begin{aligned}
&D_{\text{KL}}(\tilde{p}(\boldsymbol{\psi}, \boldsymbol{\phi}) p(\boldsymbol{\tau} \vert \boldsymbol{\psi}, \boldsymbol{\phi}) \| \tilde{p}(\boldsymbol{\tau}) \boldsymbol{q}(\boldsymbol{\psi}, \boldsymbol{\phi} \vert \boldsymbol{\tau})) \\
=& -\mathbb{E}_{\tilde{p}(\boldsymbol{\psi}, \boldsymbol{\phi}) p(\boldsymbol{\tau} \vert \boldsymbol{\psi}, \boldsymbol{\phi})} \left[ \log \boldsymbol{q}(\boldsymbol{\psi}, \boldsymbol{\phi} \vert \boldsymbol{\tau})\right] + \mathtt{const}.
\end{aligned}
\end{equation}

A KL divergence is non-negative and equals zero if and only if its two arguments coincide almost everywhere. Since $ \boldsymbol{q} $ ranges over all conditional densities, the value zero is attainable, so the KL divergence is minimized if and only if
\begin{equation}
\tilde{p}(\boldsymbol{\psi}, \boldsymbol{\phi}) p(\boldsymbol{\tau} \vert \boldsymbol{\psi}, \boldsymbol{\phi}) = \tilde{p}(\boldsymbol{\tau}) \boldsymbol{q}(\boldsymbol{\psi}, \boldsymbol{\phi} \vert \boldsymbol{\tau})
\end{equation}
almost everywhere, that is, if and only if
\begin{equation}
\begin{aligned}
\boldsymbol{q}(\boldsymbol{\psi}, \boldsymbol{\phi} \vert \boldsymbol{\tau}) &= \frac{\tilde{p}(\boldsymbol{\psi}, \boldsymbol{\phi}) p(\boldsymbol{\tau} \vert \boldsymbol{\psi}, \boldsymbol{\phi})}{\tilde{p}(\boldsymbol{\tau})} \\
&= \frac{\tilde{p}(\boldsymbol{\psi}, \boldsymbol{\phi})}{\tilde{p}(\boldsymbol{\tau})} \cdot \frac{p(\boldsymbol{\psi}, \boldsymbol{\phi} \vert \boldsymbol{\tau}) p(\boldsymbol{\tau})}{p(\boldsymbol{\psi}, \boldsymbol{\phi})} \\
&\propto \frac{\tilde{p}(\boldsymbol{\psi}, \boldsymbol{\phi})}{p(\boldsymbol{\psi}, \boldsymbol{\phi})} p(\boldsymbol{\psi}, \boldsymbol{\phi} \vert \boldsymbol{\tau})
\end{aligned}
\end{equation}
holds almost everywhere.
\end{proof}

\begin{corollary}
When the proposal prior coincides with the true prior, $ \tilde{p}(\boldsymbol{\psi}, \boldsymbol{\phi}) = p(\boldsymbol{\psi}, \boldsymbol{\phi}) $, the learned density $ \boldsymbol{q}(\boldsymbol{\psi}, \boldsymbol{\phi} \vert \boldsymbol{\tau}) $ recovers the true posterior:
\begin{equation}
\begin{aligned}
\boldsymbol{q}(\boldsymbol{\psi}, \boldsymbol{\phi} \vert \boldsymbol{\tau}) = \frac{\tilde{p}(\boldsymbol{\psi}, \boldsymbol{\phi}) p(\boldsymbol{\tau} \vert \boldsymbol{\psi}, \boldsymbol{\phi})}{\iint \tilde{p}(\boldsymbol{\psi}, \boldsymbol{\phi}) p(\boldsymbol{\tau} \vert \boldsymbol{\psi}, \boldsymbol{\phi}) \mathrm{d}\boldsymbol{\psi} \mathrm{d}\boldsymbol{\phi}}.
\end{aligned}
\end{equation}
\end{corollary}

Proposition~\ref{prop:proposal} tells us what sequential NFC learns when trained on data drawn from the simulation-pretrained proposal. Writing $ \tilde{p}(\boldsymbol{\psi}, \boldsymbol{\phi} \vert \boldsymbol{\tau}^{\mathrm{real}}) $ for the learned density, we relate it to the true posterior through
\begin{equation}
\tilde{p}(\boldsymbol{\psi}, \boldsymbol{\phi}\vert \boldsymbol{\tau}^{\mathrm{real}}) = p(\boldsymbol{\psi}, \boldsymbol{\phi}\vert \boldsymbol{\tau}^{\mathrm{real}}) \frac{\tilde{p}(\boldsymbol{\psi}, \boldsymbol{\phi})}{p(\boldsymbol{\psi}, \boldsymbol{\phi})} \frac{p(\boldsymbol{\tau}^{\mathrm{real}})}{\tilde{p}(\boldsymbol{\tau}^{\mathrm{real}})},
\end{equation}
i.e., the true posterior reweighted by the proposal-to-prior ratio. We use this learned density directly for fine-tuning, since the proposal concentrates it on parameters that already explain the observed execution. Each online update trains on the newly simulated trajectories only, warm-started from the current network parameters, and is therefore much faster than simulation pretraining.

The proposal prior $ \tilde{p}(\cdot) $ is where prior knowledge enters NFC. Semantic priors from experts or foundation models~\citep{ma2024dreureka, Ewen-RSS-24} inform specific quantities, such as ground friction inferred from vision and touch, and extend to further quantities as the corresponding real-world data becomes available. We instead seed the proposal from simulation, where trajectories can be generated across the entire parameter space without real-world costs. The two sources of knowledge are compatible, since a semantic prior can enter as an additional factor over the parameters it covers, and we adopt the simulation-pretrained proposal throughout this work.

\subsection{Latent Neural Fidelity Calibration}\label{sec:neural-dr}

While sequential NFC enables efficient online adaptation, extending it to high-dimensional observations and inferred parameters remains computationally intensive for real-time policy improvement. Calibrating low-dimensional simulator parameters---such as mass, damping, or friction---is relatively fast. However, real-world robotics systems often rely on perception modules such as depth cameras~\citep{zhuang2023robot, David24anymalparkour, zhuang2024humanoid}, introducing significant overhead for training and sampling in such high-dimensional spaces.

\begin{figure}[t]
\vspace*{0.04in}
\centering
{\setlength{\fboxsep}{0pt}\setlength{\fboxrule}{0.4pt}\fbox{\includegraphics[width=\dimexpr\linewidth-2\fboxrule\relax]{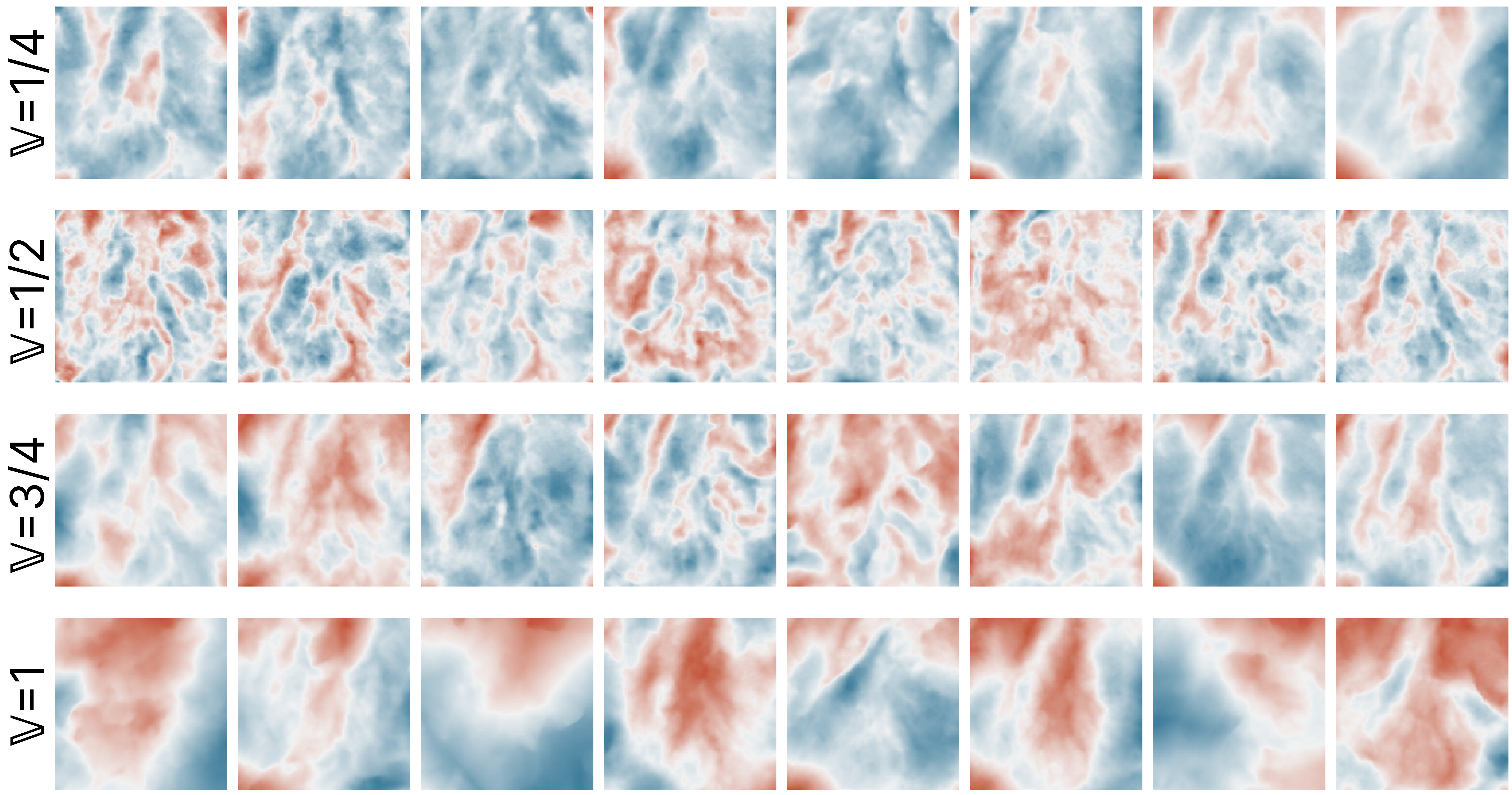}}}
\caption{Neural Fidelity Calibration enables sampling of high-dimensional residual environment geometries by accounting for perception uncertainty while preserving realism. The residual captures the difference between the onboard perceived and ground-truth environment. As the sampling variance $\mathbb{V} \in (0, 1]$ increases ($\mathbb{V} \to 0$ eliminates randomness, while $\mathbb{V} = 1$ samples from the full learned posterior), the generated samples exhibit greater diversity in appearance.}
\label{fig:FM-sample-var}
\end{figure}

To address this, we propose latent NFC, which infers the residual fidelity through a compact latent rather than in the raw spaces of $\Delta \boldsymbol{s}$ and $\Delta \boldsymbol{e} \in \mathbb{R}^{R \times C}$. Latent NFC is built upon sequential NFC. We denote the execution data over a history of length $H$ as
\begin{equation}
\boldsymbol{\tau} = \{[s_t, a_t]_{t=0}^{H},\ \boldsymbol{e}_H\},
\end{equation}
where $\boldsymbol{e}_H$ is a temporally fused perception map (e.g., a signed distance field); this representation generalizes across sensor modalities. An encoder $\mathcal{H}(\cdot)$ compresses $\boldsymbol{\tau}$ into a low-dimensional embedding. Specifically, we use a causal temporal convolutional network (TCN)~\citep{BaiTCN2018} for the state--action history and an MLP for the perception map, and concatenate their outputs. In simulation, two versions of every trajectory are available: the trajectory $\hat{\boldsymbol{\tau}}$ that the robot would actually observe, whose states are corrupted by sensor noise, and the clean, privileged trajectory $\boldsymbol{\tau}$ with true states and true terrain. Their embedding difference defines the \emph{latent residual fidelity}:
\begin{equation}\label{eqn:latent-residual}
\boldsymbol{\phi}_{z} = \mathcal{H}(\boldsymbol{\tau}) - \mathcal{H}(\hat{\boldsymbol{\tau}}),
\end{equation}
the latent counterpart of $\boldsymbol{\phi} = \{\Delta \boldsymbol{s}, \Delta \boldsymbol{e}\}$. 
The estimator conditions on $\mathcal{H}(\hat{\boldsymbol{\tau}})$ and infers the joint $\boldsymbol{q}(\boldsymbol{\psi}, \boldsymbol{\phi}_{z} \vert \cdot)$, which factorizes by construction: a flow matching model over the physical parameters in their native space, and a residual head that predicts the latent residual together with its scale from the same embedding. An MLP decoder $\mathcal{Z}(\cdot)$ maps $\boldsymbol{\phi}_{z}$ to the residual fidelity parameters $\boldsymbol{\phi}$. Together, the decoded $(\boldsymbol{\psi}, \boldsymbol{\phi})$ define the distribution of simulated domains used for policy fine-tuning. The reconstruction loss
\begin{equation}\label{eqn:recon-loss-func}
\mathcal{L}_{\mathrm{Recon}}(\boldsymbol{\tau}, \hat{\boldsymbol{\tau}} \vert \mathcal{H}, \mathcal{Z}) = \big\| \mathcal{Z}\big(\mathcal{H}(\boldsymbol{\tau}) - \mathcal{H}(\hat{\boldsymbol{\tau}})\big) - \boldsymbol{\phi} \big\|_2^2
\end{equation}
ensures that the latent retains the information needed to rebuild the residual dynamics and environment. We jointly optimize the velocity field, encoder, and decoder:
\begin{equation}\label{eqn:joint-loss-func}
\begin{aligned}
\min_{\boldsymbol{\theta}, \mathcal{H}, \mathcal{Z}} \frac{1}{N} \sum_{n} -\log \boldsymbol{q}_{\boldsymbol{\theta}}\big(\boldsymbol{\psi}_{n}, \boldsymbol{\phi}_{z,n} \,\big\vert\, \mathcal{H}(\hat{\boldsymbol{\tau}}_n)\big) \\+ \alpha \mathcal{L}_{\mathrm{Recon}}(\boldsymbol{\tau}_{n}, \hat{\boldsymbol{\tau}}_{n} \vert \mathcal{H}, \mathcal{Z}),
\end{aligned}
\end{equation}
where $\alpha$ controls the reconstruction regularization, and the two terms are optimized with the alternating schedule of Alg.~\ref{alg:nfc-pretrain}: we first update $\boldsymbol{q}_{\boldsymbol{\theta}}$ and $\mathcal{H}$, then freeze them and update $\mathcal{Z}$.

This joint objective encourages $ \boldsymbol{\phi}_{z} $ to encode features that are both posterior-informative and reconstructive. The alternating schedule is reminiscent of domain-adversarial training~\citep{ganin2016domain}; the reconstruction term plays a role analogous to privileged-information distillation in teacher--student sim-to-real transfer~\citep{he2024bridging}, ensuring the compressed embedding retains the information needed for inference under noisy, high-dimensional observations. Pseudocode is given in Alg.~\ref{alg:nfc-pretrain}.

\begin{algorithm}[t]
\caption{Latent NFC Learning}\label{alg:nfc-pretrain}
\SetKwInOut{KwIn}{Input}
\SetKwInOut{KwOut}{Output}
\SetKwInOut{KwInit}{Initialize}

\KwIn{Simulated trajectories $\boldsymbol{\tau}$, physical parameters $\boldsymbol{\psi}$, residual fidelity $\boldsymbol{\phi}$, encoder $\mathcal{H}$, decoder $\mathcal{Z}$, NFC model $\boldsymbol{q}(\boldsymbol{\psi}, \boldsymbol{\phi}_{z} \vert \cdot)$, optimizer $\mathtt{opt}(\cdot)$, threshold $\epsilon$}
\KwOut{Pretrained NFC model $\boldsymbol{q}$, encoder $\mathcal{H}$, decoder $\mathcal{Z}$}
\KwInit{Domain randomization parameters for states and perception maps}

\tcc{\textcolor{blue}{NFC Pretraining in Simulation}}
\For{$i \gets 1$ \KwTo max iterations}{
    \ForEach{trajectory $\boldsymbol{\tau}$ in \texttt{batch}($\mathcal{D}_{\text{sim}}$)}{
        \tcc{Domain perturbation}

        Sample noisy trajectory $\hat{\boldsymbol{\tau}} \sim \mathcal{P}(\boldsymbol{\tau})$\;
        $\boldsymbol{\phi}_{z} \gets \mathcal{H}(\boldsymbol{\tau}) - \mathcal{H}(\hat{\boldsymbol{\tau}})$ \tcp*{Eq.~\eqref{eqn:latent-residual}}
        
        \tcc{Update NFC and encoder}

        $\mathcal{L}_{\text{NFC}} \gets -\log \boldsymbol{q}(\boldsymbol{\psi}, \boldsymbol{\phi}_{z} \vert \mathcal{H}(\hat{\boldsymbol{\tau}})) + \alpha \mathcal{L}_{\mathrm{Recon}}(\boldsymbol{\tau}, \hat{\boldsymbol{\tau}} \vert \mathcal{H}, \mathcal{Z})$\;
        Update $\boldsymbol{q}, \mathcal{H} \gets \mathtt{opt}(\mathcal{L}_{\text{NFC}})$\;
        
        \tcc{Freeze encoder/NFC and update decoder}
        $\mathcal{L}_{\mathrm{Recon}} \gets \mathcal{L}_{\mathrm{Recon}}(\boldsymbol{\tau}, \hat{\boldsymbol{\tau}} \vert \mathcal{H}, \mathcal{Z})$\;
        Update $\mathcal{Z} \gets \mathtt{opt}(\mathcal{L}_{\mathrm{Recon}})$\;
    }
    \If{$\mathrm{mean}(\mathcal{L}_{\text{NFC}}) < \epsilon$ \textbf{and} $\mathrm{mean}(\mathcal{L}_{\mathrm{Recon}}) < \epsilon$}{
        \textbf{break}\;
    }
}
\end{algorithm}

\begin{running-example}
\emph{
Revisiting the unstructured terrain navigation task, we demonstrate the application of latent NFC to model residual fidelity. Fig.~\ref{fig:FM-sample-var} shows sampled residual elevation maps $ \Delta \boldsymbol{e} \in \mathbb{R}^{128 \times 128} $, where each pixel captures the elevation difference between the ground-truth terrain and the robot's perceived environment. These samples illustrate how latent NFC refines environment reconstructions by inferring sensor-perception errors. Concretely, we draw $N$ joint samples $ (\boldsymbol{\psi}, \boldsymbol{\phi}_{z}) $ with $ \boldsymbol{\psi} \in \mathbb{R}^{13} $; the decoder $ \mathcal{Z} $ maps each $ \boldsymbol{\phi}_{z} $ to a residual state offset ($ \mathbb{R}^{13} $) and a residual environment ($ \mathbb{R}^{128 \times 128} $).
}
\end{running-example}

While generative models such as diffusion have been successfully applied to generate large-scale geometries~\citep{jia2024cluttergen, liu2024pyramid}, our focus is on residual fidelity modeling for robot perception. Rather than generating the entire environment $ \boldsymbol{e} $, we instead model the residual $ \Delta \boldsymbol{e} = \boldsymbol{e} - \hat{\boldsymbol{e}} $, which represents the discrepancy between the ground-truth and perceived environments. This design is motivated by recent findings in image restoration~\citep{Yue2025Restoration}, where learning residuals improves both efficiency and quality of reconstructions compared to modeling full images.

In essence, our residual fidelity modeling offers two key advantages: (1) continuous-time generative estimators outperform conventional neural networks for simulation-based inference in high dimensions~\citep{sharrock2024sequential, wildberger2023flow}; (2) learning in the residual space enhances flow matching inference accuracy and training performance. By jointly calibrating simulator parameters and residual fidelity domains, our framework enables fast adaptation, realistic simulation, and robust sim-to-real policy adaptation.

\subsection{Policy Optimization with Hallucinated Randomness}\label{sec:hallucinate}

Given posterior samples from NFC, the direct approach is domain randomization with the posterior in place of hand-designed ranges: sample simulator domains from $\boldsymbol{q}$ and maximize expected return across them,
\begin{equation}\label{eqn:greedy-policy-opt}
\boldsymbol{\pi}^{*} = \operatorname*{argmax}_{\boldsymbol{\pi}} \; \mathbb{E}_{(\boldsymbol{\psi}, \boldsymbol{\phi}) \sim \boldsymbol{q}(\cdot \mid \hat{\boldsymbol{\tau}})} \, \mathbb{E}_{\boldsymbol{\tau} \sim P^{\boldsymbol{\pi}}_{\boldsymbol{\psi}, \boldsymbol{\phi}}} \left[ R(\boldsymbol{\tau}) \right],
\end{equation}
where $P^{\boldsymbol{\pi}}_{\boldsymbol{\psi}, \boldsymbol{\phi}}$ is the rollout distribution of $\boldsymbol{\pi}$ in the simulator under the sampled context. This mirrors the exploitation step of model-based RL~\citep{deisenroth2011pilco, chua2018deep}, with the calibrated simulator as the model. When the posterior is sharp, training on its samples tailors the policy to the inferred real-world dynamics.

However, when the posterior is highly uncertain, directly sampling parameters may impair policy learning. For instance, fine-tuning in very low-friction environments may cause the robot to slide uncontrollably, whereas high friction could result in the robot getting stuck. In both cases, the robot struggles to obtain reward signals within a short time, hindering effective online adaptation. To mitigate this, we introduce a hallucination policy $ \boldsymbol{\pi}_{\mathfrak{h}} $ that never acts on the robot; instead, it controls how far each fine-tuning domain may deviate from the center of the NFC posterior. 

Formally, let $ \boldsymbol{x} = [\boldsymbol{\psi}, \boldsymbol{\phi}_{z}] \in \mathbb{R}^{D} $ denote a joint NFC sample in latent form (Sec.~\ref{sec:neural-dr}), with $ D = \vert \boldsymbol{\psi} \vert + \vert \boldsymbol{\phi}_{z} \vert $, and let $ \boldsymbol{\mu} $ be the mean of the $ N $ samples drawn in the current adaptation round. Conditioned on the embedding of the observed history, the hallucination policy samples one gate vector per parallel environment and rollout, $ \boldsymbol{g} \sim \boldsymbol{\pi}_{\mathfrak{h}}(\cdot \mid \mathcal{H}(\hat{\boldsymbol{\tau}})) $ with $ \boldsymbol{g} \in (0, 1]^{D} $, and each simulated domain is
\begin{equation}\label{eqn:hallucinate-domain}
\tilde{\boldsymbol{x}} = \boldsymbol{\mu} + \boldsymbol{g} \odot (\boldsymbol{x} - \boldsymbol{\mu}), \qquad \boldsymbol{x} \sim \boldsymbol{q}(\cdot \mid \hat{\boldsymbol{\tau}}),
\end{equation}
which $ \mathcal{Z} $ decodes into $ (\tilde{\boldsymbol{\psi}}, \tilde{\boldsymbol{\phi}}) $ to define the simulated domain. Setting $ \boldsymbol{g} = \boldsymbol{1} $ recovers the NFC posterior exactly, and $ \boldsymbol{g} \to \boldsymbol{0} $ collapses all domains to the posterior mean. Because the gate is per-coordinate, the agent can retain the posterior variance of some coordinates (e.g., wheel friction) while shrinking others (e.g., the residual-terrain latent). The gate is initialized at $ \boldsymbol{1} $ and smoothed with an exponential moving average. Since it acts on samples rather than on distribution parameters, it applies to whatever posterior the flow matching model represents without a parametric assumption.

We refer to $ \boldsymbol{\pi}_{\mathfrak{h}} $ as the hallucination policy because it allows the agent to ``hallucinate'' slightly more favorable---but plausible---dynamics. This mechanism draws from the principle of optimism under uncertainty~\citep{JMLR:v11:jaksch10a}, which encourages optimistic exploration and has shown success in model-based RL (MBRL) \citep{curi2020efficient}. Unlike MBRL methods that learn the robot's dynamics model, our approach treats the simulator as the model and learns its parameters. The policy optimization objective becomes:
\begin{equation}\label{eqn:final-policy-opt}
\boldsymbol{\pi}^{*}, \boldsymbol{\pi}_{\mathfrak{h}}^{*} = \operatorname*{argmax}_{\boldsymbol{\pi}, \boldsymbol{\pi}_{\mathfrak{h}}} \;
\mathbb{E}_{\boldsymbol{x} \sim \boldsymbol{q}(\cdot \mid \hat{\boldsymbol{\tau}})} \,
\mathbb{E}_{\boldsymbol{g} \sim \boldsymbol{\pi}_{\mathfrak{h}}(\cdot \mid \mathcal{H}(\hat{\boldsymbol{\tau}}))} \,
\mathbb{E}_{\boldsymbol{\tau} \sim P^{\boldsymbol{\pi}}_{\tilde{\boldsymbol{x}}}} \left[ R(\boldsymbol{\tau}) \right],
\end{equation}
where $ \tilde{\boldsymbol{x}} $ is given by Eq.~\eqref{eqn:hallucinate-domain}. We solve this joint optimization using standard policy gradient methods. In all experiments, we use PPO~\citep{schulman2017proximal} policies for the robot control policy $\boldsymbol{\pi}$ and the hallucination policy $\boldsymbol{\pi}_{\mathfrak{h}}$.

\section{Informative Sim-to-Real Adaptation}\label{sec:adaptation}
Building on the Neural Fidelity Calibration framework, we now introduce a real-world deployment strategy that enables informative policy adaptation for mission-critical robotics. The central idea is to prioritize online adaptation under anomalous conditions where the pretrained policy is likely to fail, while leveraging offline refinement in nominal scenarios to improve robustness over time.

Under nominal conditions (e.g., a compact robot navigating structured terrain), the simulation-pretrained policy performs well. We collect real-world execution data in these cases and use it offline to improve both policy and NFC. In contrast, for failure-prone scenarios (e.g., a damaged robot on slippery terrain), we perform online adaptation, correcting policy behavior immediately through NFC. To enable this, we introduce (1) an anomaly detection module to trigger online adaptation, and (2) a runtime protocol that governs the interaction between adaptation phases.

\subsection{Anomaly Detection}\label{sec:anomaly-detect}
We define anomalies as deviations between expected and observed robot behavior that may compromise the pretrained policy. We detect these deviations by comparing a nominal reference window $\boldsymbol{\tau}^{\mathfrak{c}}$ with a target window $\boldsymbol{\tau}^{\mathfrak{t}}$, following the contrastive-window formulation for time-series anomaly detection~\citep{darban2025carla}. The comparison itself is a classifier two-sample test~\citep{lopez2016revisiting}: a discriminator is trained to distinguish pairs drawn from the same regime from pairs drawn from different regimes, and its score is thresholded at deployment. We instantiate the discriminator as a Siamese network~\citep{koch2015siamese}, whose encoder is the causal TCN~\citep{BaiTCN2018} used by NFC. The key idea is summarized below.

Let $\mathcal{H}(\boldsymbol{\tau})$ denote the flattened temporal activation map of the detector's TCN, applied with shared weights to both windows. Following~\citet{koch2015siamese}, the anomaly logit is a single affine classifier on the elementwise absolute difference,
\begin{equation}\label{eq:anomaly-score}
\ell_{\mathrm{A}} = \boldsymbol{w}^{\top} \left| \mathcal{H}(\boldsymbol{\tau}^{\mathfrak{c}}) - \mathcal{H}(\boldsymbol{\tau}^{\mathfrak{t}}) \right| + b,
\end{equation}
which retains differences in feature magnitude without a projection MLP, unit normalization, cosine margin, or temperature. For a pair label $y \in \{0,1\}$, encoder and classifier are trained jointly with binary cross-entropy,
\begin{equation}\label{eq:anomaly-loss}
\mathcal{L}_{\mathrm{Anomaly}} = -y \log \sigma(\ell_{\mathrm{A}}) - (1-y) \log\left[1 - \sigma(\ell_{\mathrm{A}})\right],
\end{equation}
where $\sigma$ is the sigmoid. What distinguishes our instantiation is the source of the labels: training pairs are separate simulated episodes, receiving $y = 0$ when the physical and residual parameters are retained and $y = 1$ when the target episode's parameters are independently redrawn from the prior. The detector therefore needs no taxonomy of possible faults, only the parameter supports NFC already assumes (Assumption~\ref{asm:support}). At deployment a target window is flagged when $\ell_{\mathrm{A}} > \eta$, with $\eta$ the \num{99}th percentile of scores on held-out nominal pairs.

\subsection{Informative Sim-to-Real Adaptation with NFC}\label{sec:adapt-nfc}

\begin{algorithm}
\vspace*{0.04in}
\caption{Informative Sim-to-Real Adaptation with NFC}
\label{alg:nfc}
\SetKwInOut{KwIn}{Input}
\SetKwInOut{KwOut}{Output}
\SetKwInOut{KwInit}{Initialize}
\KwIn{Initial policy $ \boldsymbol{\pi} $, NFC $ \boldsymbol{q}(\boldsymbol{\psi}, \boldsymbol{\phi}_{z} \vert \cdot) $ with encoder $ \mathcal{H} $ and decoder $ \mathcal{Z} $, anomaly detector, and hallucination policy $ \boldsymbol{\pi}_{\mathfrak{h}} $}
\KwOut{Optimized policy $ \boldsymbol{\pi}^{*} $, Dataset $ \mathcal{T} $}
\tcc{\textcolor{blue}{Offline Pretraining Phase}}
\KwInit{The policy $ \boldsymbol{\pi} $, NFC $ \boldsymbol{q} $, and anomaly detector through simulation domain randomization training (Alg.~\ref{alg:nfc-pretrain})}

\tcc{\textcolor{blue}{Online Deployment Phase}}
\While{Mission not Complete}{
    \tcc{\small Trajectory with History $H$}
    Execute trajectory $ \boldsymbol{\tau}_{t} $
    
    \tcc{Anomaly test, Eq.~\eqref{eq:anomaly-score}}
    $\ell_{\mathrm{A}} \gets \boldsymbol{w}^{\top} | \mathcal{H}(\boldsymbol{\tau}^{\mathfrak{c}}) - \mathcal{H}(\boldsymbol{\tau}_{t}) | + b$\;
    \If{$\ell_{\mathrm{A}} > \eta$}{
        \tcc{\texttt{NFC}: draw $N$ samples}

        $ \{\boldsymbol{x}_n = [\boldsymbol{\psi}_n, \boldsymbol{\phi}_{z,n}]\}_{n=1}^{N} \sim \boldsymbol{q}(\cdot \vert \mathcal{H}(\boldsymbol{\tau}_t)) $, \quad $ \boldsymbol{\mu} \leftarrow \frac{1}{N}\sum_n \boldsymbol{x}_n $ \tcp*{Eq.~\eqref{eqn:neural-posterior}}

        \tcc{\texttt{Hallucination}: gate the spread}
        $\boldsymbol{g}_n \sim \boldsymbol{\pi}_{\mathfrak{h}}(\cdot \mid \mathcal{H}(\boldsymbol{\tau}_t))$,       
        $\tilde{\boldsymbol{x}}_n \leftarrow \boldsymbol{\mu} + \boldsymbol{g}_n \odot (\boldsymbol{x}_n - \boldsymbol{\mu})$ \tcp*{Eq.~\eqref{eqn:hallucinate-domain}}

        \tcc{Decode and reset $N$ parallel simulators}

        $ \tilde{\boldsymbol{\phi}}_n \leftarrow \mathcal{Z}(\tilde{\boldsymbol{\phi}}_{z,n}) $, \quad $ \boldsymbol{e}_n \leftarrow \hat{\boldsymbol{e}} + \Delta \boldsymbol{e}_n $

        \tcc{\texttt{Improve Policy} Eq.~\eqref{eqn:final-policy-opt}}
        $ \boldsymbol{\pi}, \boldsymbol{\pi}_{\mathfrak{h}} \leftarrow \texttt{Policy-Opt}(\boldsymbol{\pi}, \boldsymbol{\pi}_{\mathfrak{h}}, \{\tilde{\boldsymbol{\psi}}_n, \tilde{\boldsymbol{\phi}}_n\}_{n=1}^{N})$; collect rollouts $ \{\boldsymbol{\tau}^{\mathrm{sim}}_n\} $

        \tcc{\texttt{Sequential NFC} (Prop.~\ref{prop:proposal}): the gated samples act as proposal}
        $ \boldsymbol{q} \leftarrow $ train on $ \{(\tilde{\boldsymbol{x}}_n, \boldsymbol{\tau}^{\mathrm{sim}}_n)\}_{n=1}^{N} $
    }
    
    \tcc{\texttt{Update Dataset}}
    $ \mathcal{T} = \mathcal{T} \cup \boldsymbol{\tau}_{t}$, $ t \leftarrow t + H $ 
}

\tcc{\textcolor{blue}{Post-Mission Offline Phase}}
Refine Policy, NFC, and Anomaly Detector with $ \mathcal{T} $
\end{algorithm}

Alg.~\ref{alg:nfc} summarizes the full procedure. Offline pretraining trains the policy, NFC, and anomaly detector in simulation under domain randomization~\citep{tobin2017domain}. During deployment, each detected anomaly triggers one adaptation round: NFC infers the posterior from the most recent window of $ H $ steps, the hallucination policy modulates its uncertainty, and the policy is fine-tuned in the resampled simulator.
Because inference is conditioned on the most recent window, NFC can infer different domain distributions at successive triggers. Recovery of an earlier behavior additionally depends on the policy learner, since fine-tuning updates persist across rounds~\citep{schulman2017proximal}. After the mission, the accumulated dataset $ \mathcal{T} $ refines NFC, the detector, and the policy, enabling continual improvement across deployments.

\section{Simulation Experiment}\label{sec:sim-exp}

\begin{table*}[t]
\small
\centering
\setlength{\tabcolsep}{4.5pt}
\renewcommand{\arraystretch}{1.08}
\caption{
Physical parameter and residual fidelity priors per robot embodiment. Intervals $[a, b]$ denote uniform distributions $\mathcal{U}[a, b]$.
}
\label{tab:sim-physic-params}
\begin{tabular*}{\textwidth}{@{\extracolsep{\fill}} l ccccc @{}}
\toprule
\textbf{Property} & \texttt{Ant} & \texttt{Quadruped} & \texttt{Humanoid} & \texttt{Quadcopter} & \texttt{Jackal} \\
\midrule
\multicolumn{6}{l}{\textit{Physical Parameters} $\boldsymbol{\psi}$ (\texttt{Ant}: $17$, \texttt{Quadruped}: $25$, \texttt{Humanoid}: $39$, \texttt{Quadcopter}: $9$, \texttt{Jackal}: $13$)} \\
\quad Mass & $\times\,[0.01, 5]^9$ & $\times\,[0.01, 5]^{13}$ & $\times\,[0.1, 2]^{20}$ & $\times\,[0.5, 5]^5$ & $+\,[0, 5]^5$ \\
\quad Stiffness / Rotor gain & $+\,[0, 20]^8$ & $+\,[0, 20]^{12}$ & $+\,[0, 10]^{19}$ & $+\,[0, 20]^4$ & $-$ \\
\quad Damping & $-$ & $-$ & $-$ & $-$ & $\times\,[0, 1]^4$ \\
\quad Friction & $-$ & $-$ & $-$ & $-$ & $=\,[0.1, 1.3]^4$ \\
\midrule
\multicolumn{6}{l}{\textit{Residual Fidelity} $\boldsymbol{\phi} = \{\Delta\boldsymbol{s}, \Delta\boldsymbol{e}\}$ (\texttt{Ant}: $254$, \texttt{Quadruped}: $252$, \texttt{Humanoid}: $252$, \texttt{Quadcopter}: $250$, \texttt{Jackal}: $252$)} \\
\quad Base height & $+\,[0, 0.01]^1$ & $-$ & $-$ & $-$ & $-$ \\
\quad Linear velocity & $+\,[0, 0.1]^3$ & $+\,[0, 0.1]^3$ & $+\,[0, 0.1]^3$ & $+\,[0, 0.1]^3$ & $+\,[0, 0.1]^3$ \\
\quad Angular velocity & $+\,[0, 0.2]^3$ & $+\,[0, 0.2]^3$ & $+\,[0, 0.2]^3$ & $+\,[0, 0.2]^3$ & $+\,[0, 0.2]^3$ \\
\quad Projected gravity & $+\,[0, 0.01]^3$ & $+\,[0, 0.01]^3$ & $+\,[0, 0.01]^3$ & $+\,[0, 0.01]^3$ & $+\,[0, 0.01]^3$ \\
\quad Joint position & $+\,[0, 0.01]^8$ & $+\,[0, 0.01]^{12}$ & $+\,[0, 0.01]^{19}$ & $-$ & $-$ \\
\quad Joint velocity & $+\,[0, 0.1]^8$ & $+\,[0, 0.1]^{12}$ & $+\,[0, 0.1]^{19}$ & $-$ & $-$ \\
\quad Goal position & $-$ & $-$ & $-$ & $+\,[0, 0.01]^3$ & $+\,[0, 0.1]^3$ \\
\quad Heading & $-$ & $-$ & $-$ & $-$ & $+\,[0, 0.01]^1$ \\
\quad Terrain elevation & $+\,[-0.1, 0.1]^{128^2}$ & $+\,[-0.1, 0.1]^{128^2}$ & $+\,[-0.1, 0.1]^{128^2}$ & $-$ & $+\,[-0.1, 0.1]^{128^2}$ \\
\quad Depth map & $-$ & $-$ & $-$ & $+\,[-0.03, 0.03]^{6360}$ & $-$ \\
\bottomrule
\end{tabular*}
\vspace{2pt}
\raggedright \footnotesize 
Superscripts give the perturbed group dimension. NFC infers one bias and one noise scale per group, so $\Delta\boldsymbol{s}$ has twice the number of groups; $\Delta\boldsymbol{e}$ is inferred as one map-level bias and scale together with \num{120} latent corrections and \num{120} uncertainties, decoded to the grid. Prefixes denote nominal scaling ($\times$), additive shift ($+$), and absolute value ($=$). For $\Delta\boldsymbol{s}$, $+[0, u]$ indicates episode bias from $[-u, u]$ and Gaussian noise scale from $[0, u]$. For $\Delta\boldsymbol{e}$, ranges denote additive bounds.
\end{table*}

To validate our approach before real-world deployment, we conduct sim-to-sim experiments, focusing on three core capabilities: anomaly detection, NFC accuracy, and policy improvement via NFC. These evaluations serve to isolate and quantify the contributions of each component under controlled conditions. Building on the NFC framework described in previous sections, we aim to answer the following research questions:
\begin{enumerate}[label=\textbf{Q\arabic*.}, leftmargin=*]
\item \textbf{Anomaly Detection}: Does the detector trigger adaptation reliably across platforms and terrains, and which feature representation achieves the best true positive rate?
\item \textbf{NFC Performance}: How precisely can NFC infer high-dimensional simulator parameters and residual fidelity representations, relative to state-of-the-art baselines?
\item \textbf{Policy Improvement}: To what extent does NFC enhance sim-to-real transfer through informed fine-tuning, and what is the impact of the residual fidelity module?
\end{enumerate}

\subsection{Sim-to-Sim Experiment Settings}
\label{sec:sim-settings}

\textbf{Robots and Environments.}
We evaluate on five representative robotic platforms---\texttt{Ant}, \texttt{Quadruped} (Unitree Go1), \texttt{Humanoid} (Unitree H1), \texttt{Quadcopter} (Ingenuity), and \texttt{Jackal} (Clearpath)---spanning legged, aerial, and wheeled locomotion. Policies and NFC are pretrained in Genesis World~\citep{Genesis} and deployed in MuJoCo~\citep{todorov2012mujoco}, whose differing contact and integration models provide the sim-to-sim gap; we additionally inject Gaussian noise in MuJoCo. Throughout this section, MuJoCo plays the role of the real world, so ``real-world'' data and trajectories refer to MuJoCo rollouts. Each robot carries a simulated RealSense D435 from which elevation maps are built~\citep{RAL18-elemap}, except \texttt{Quadcopter}, which perceives depth images directly. We use two environment types: \emph{Flat}, an obstacle-free world, and \emph{Rough}, with randomized elevation maps for ground robots~\citep{yu2024adaptive} and randomized 3D obstacle fields for \texttt{Quadcopter}~\citep{kulkarni2023aerialgymisaac}.

\textbf{Task Interface.}
\texttt{Jackal} performs goal-reaching navigation~\citep{yu2024adaptive}. The control policy observes its proprioceptive state, the ego-centric elevation map, and the current goal in the robot frame, and outputs linear and angular velocity commands. In simulation, one goal is sampled per episode at a distance of \SIrange{3}{6}{\meter}. In the physical experiments of Sec.~\ref{sec:real-world}, the robot follows a sequence of waypoints along a figure-\num{8} course, advancing once within \SI{0.25}{\meter}. NFC and the anomaly detector consume the TCN embedding of the trajectory window $\boldsymbol{\tau}_t$ of state--action features, and NFC is additionally conditioned on the fused perception map $\boldsymbol{e}_H$. \texttt{Quadruped} and \texttt{Humanoid} track velocity commands, observing proprioceptive state together with the commanded velocity; \texttt{Quadcopter} uses a target-position reward and \texttt{Ant} a progress reward toward a distant target. The legged policies output joint-position targets and the aerial policy outputs rotor thrusts.

\textbf{Implementation Details.}
Table~\ref{tab:sim-physic-params} lists, per robot, the physical parameters $\boldsymbol{\psi}$ with their priors, the residual dynamics $\Delta\boldsymbol{s}$, and the perception map corrected by the residual environment $\Delta\boldsymbol{e}$ (\SI{0.1}{\meter} resolution for elevation maps). Policies are pretrained in Genesis with PPO across \num{4096} parallel environments. The NFC pretraining set comprises at most \num{1e6} windows of $H = 50$ control steps (policy at \SI{25}{\hertz}).

The encoder $\mathcal{H}$ is a causal TCN with channel widths $[128, 32]$, kernel size \num{8}, and LeakyReLU activation. The NFC posterior estimator is parameterized by a ResNet with hidden width \num{1024}. The decoder $\mathcal{Z}$ is an MLP with five hidden layers of \num{256} units and LeakyReLU activation. The anomaly detector has the TCN encoder followed by an affine classifier applied to the absolute difference between reference and current-window embeddings. NFC and anomaly-detector training use AdamW at a learning rate of \num{1e-3}.

Regarding online time efficiency, each online policy fine-tuning round with NFC posterior sampling requires \SI{0.2}{\second} on average, achieving \SI{5}{\hertz} policy improvement. Although online NFC training takes around \SI{1.75}{\second}, it runs in parallel with the robot after each policy update. Since each policy execution window is \SI{2}{\second}, online NFC training leaves real-time performance unaffected.

\textbf{SBI Baselines and Ablations.}
We evaluate the effectiveness of our temporal convolutional network (\sbf{TCN}) and conditional flow matching estimator (\sbf{FM}) by comparing them against alternatives commonly used in robotics and simulation-based inference. To study the impact of our latent NFC (i.e., FM with TCN), we benchmark it against three established feature extraction methods under our FM:
\begin{enumerate}[label=(\arabic*), leftmargin=*]
    \item \sbf{FM-MLP}: We include a five-layer MLP (\num{512} hidden units per layer) commonly used in SBI. It also acts as our NFC ablation without latent reconstruction, since dropping the latent reconstruction loss yielded similar performance in preliminary tests.
    \item \sbf{FM-RFF}: Quasi-Monte Carlo Random Fourier Features (RFF) using Sobol sequences with \num{1024} dimensions, efficient for low-dimensional SBI~\citep{ramos2019bayessim}.
    \item \sbf{FM-FKS}: Fast Kernel Slicing (FKS) using non-equispaced FFT with \num{1024} features provides theoretical advantages over RFF in modeling non-smooth kernels~\citep{johannes2024slice}. While RFFs efficiently approximate standard shift-invariant kernels in high dimensions, FKS avoids Bochner's theorem limitations, making it theoretically superior for conditionally positive definite kernels that would otherwise be over-smoothed.
\end{enumerate}
Meanwhile, to assess the impact of our FM estimator, we further compare against representative methods:
\begin{enumerate}[label=(\arabic*), leftmargin=*, resume]
    \item \sbf{MDN-RFF}: A mixture density network (MDN) using QMC-RFF features as in BayesSim~\citep{ramos2019bayessim}, with a five-layer MLP (\num{512} hidden units per layer).
    \item \sbf{MDN-TCN}: A mixture density network using our proposed TCN encoder.
    \item \sbf{CMA-ES}: Covariance Matrix Adaptation Evolution Strategy~\citep{hansen2016cma}, which has demonstrated strong performance in estimating physical parameters such as the mass--inertia matrix~\citep{sobanbabu2025sampling}. Its objective is to minimize the mean squared error between real and simulated trajectories under identical action sequences. It is not directly applicable to residual-fidelity inference due to the dimensionality. We use Optuna~\citep{optuna_2019} with a population size equal to the number of parallel robots in simulation.
\end{enumerate}
For fairness, all kernel-based methods use the same Gaussian basis function, with \num{1024} features, quasi-equivalent to the latent-layer size of TCN and MLP. The length-scale is initialized as $ \mathtt{Softplus}(0) = 0.693 $ following \cite{xu2025standard} and optimized exclusively using simulation trajectories. Implementation details are provided in Appendix~\ref{apdx:kernel-feats}. All MDN variants employ full-covariance modeling and a \num{10}-component Gaussian mixture, using the same optimizer as our FM model.

\subsection{Anomaly Detection Results}

\begin{figure*}[t!]
\vspace*{0.08in}
\centering
{\setlength{\fboxsep}{0pt}\setlength{\fboxrule}{0.4pt}\fbox{\includegraphics[width=\dimexpr\linewidth-2\fboxrule\relax]{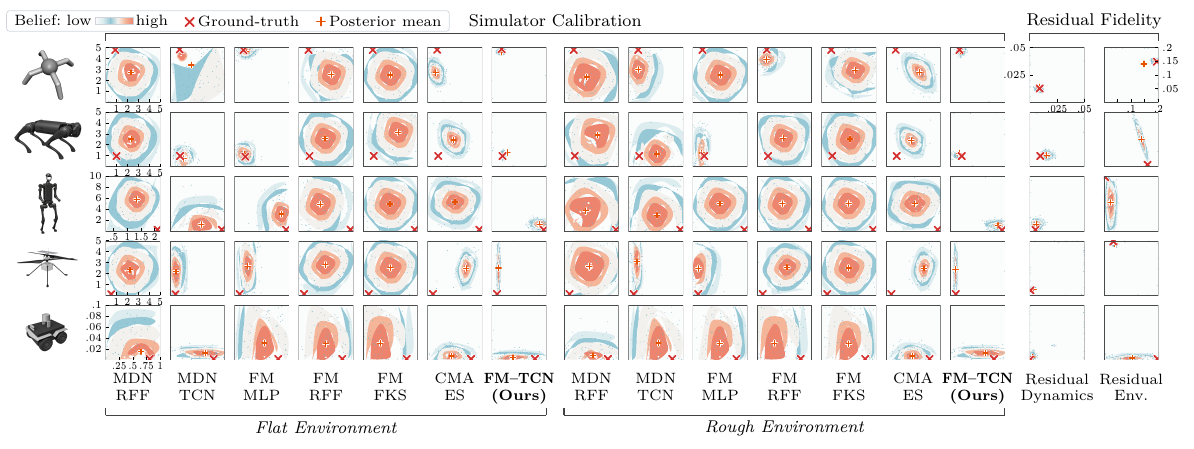}}}
\caption{
Simulator calibration posteriors across five robots for our method and competing approaches in \emph{Flat} (empty-world) and \emph{Rough} (unstructured) environments. Red crosses mark ground-truth values, orange plus symbols mark posterior means, and blue dots are posterior samples. Posterior belief increases from pale blue (low) through warm coral (high). Residual-fidelity posteriors---including residual dynamics (positions $x$ and $y$) and residual environment (height or depth in $z$)---are shown for our method only. The demonstrated parameter pairs are: [Ant] torso mass vs. left-back-foot mass; [Quadruped] right-front-shank mass vs. right-rear-shank mass; [Humanoid] torso mass vs. right-hip stiffness; [Quadcopter] first-rotor mass vs. third-rotor-arm mass; and [Jackal] front-left-wheel friction vs. damping.
}
\label{fig:sim-posterior}
\end{figure*}

\begin{table*}[t!]
\small
\centering
\setlength{\tabcolsep}{1.4pt}
\caption{
Posterior-inference accuracy (left: average log-density of the parameters) and anomaly-detection performance (right: True Positive Rate/False Positive Rate) for all inference--feature combinations. Five robots in \emph{Flat} (empty-world) and \emph{Rough} (unstructured) environments demonstrate the effect of perception on both tasks. \textbf{\textcolor{teal}{Green}} and \textbf{Bold} indicate the best and second-best results. -- denotes results unavailable due to the respective method's limitations.
}
\label{tab:anomaly-detect}
\begin{tabular*}{\textwidth}{@{\extracolsep{\fill}} ll rrrrrrrrrr rrrr @{}}
\toprule
& & \multicolumn{10}{c}{\textbf{Average Log Posterior Density of Parameters} ($\uparrow$)} & \multicolumn{4}{c}{\textbf{Anomaly Detection}} \\
\cmidrule(lr){3-12} \cmidrule(lr){13-16}
& & \multicolumn{2}{c}{\texttt{Ant}} & \multicolumn{2}{c}{\texttt{Quadruped}} & \multicolumn{2}{c}{\texttt{Humanoid}} & \multicolumn{2}{c}{\texttt{Quadcopter}} & \multicolumn{2}{c}{\texttt{Jackal}} & \multicolumn{2}{c}{TPR ($\uparrow$)} & \multicolumn{2}{c}{FPR ($\downarrow$)} \\
\cmidrule(lr){3-4} \cmidrule(lr){5-6} \cmidrule(lr){7-8} \cmidrule(lr){9-10} \cmidrule(lr){11-12} \cmidrule(lr){13-14} \cmidrule(lr){15-16}
\textbf{Inference} & \textbf{Feature} & \emph{Flat} & \emph{Rough} & \emph{Flat} & \emph{Rough} & \emph{Flat} & \emph{Rough} & \emph{Flat} & \emph{Rough} & \emph{Flat} & \emph{Rough} & \emph{Flat} & \emph{Rough} & \emph{Flat} & \emph{Rough} \\
\midrule
MDN    & QMC-RFF & $-4.88$ & $\mathbf{-0.59}$ & $-2.09$ & $-2.30$ & $-10.52$ & $-13.27$ & $-2.26$ & $-2.47$ & $\mathbf{0.64}$ & $-1.20$ & $0.70$ & $0.51$ & $0.01$ & $0.01$ \\
MDN    & TCN     & $-2.60$ & $-3.10$ & $-1.64$ & $-3.11$ & $-2.52$  & $-3.42$  & $-1.79$ & $-2.56$ & $0.22$ & $-1.17$ & \textcolor{teal}{$\mathbf{0.97}$} & \textcolor{teal}{$\mathbf{0.80}$} & $0.01$ & $0.07$ \\
\addlinespace[2pt]
FM    & MLP     & $\mathbf{-1.00}$ & $-1.09$ & $\mathbf{-1.14}$ & $\mathbf{-1.34}$ & $\mathbf{-1.63}$  & $\mathbf{-2.93}$  & $\mathbf{-1.20}$ & $\mathbf{-1.86}$ & $0.26$ & $\mathbf{-1.10}$ & $\mathbf{0.85}$ & $\mathbf{0.65}$ & $0.01$ & $0.01$ \\
FM    & QMC-RFF & $-2.79$ & $-3.03$ & $-2.17$ & $-2.48$ & $-2.61$  & $-3.33$  & $-3.00$ & $-3.64$ & $0.06$ & $-1.29$ & $0.70$ & $0.51$ & $0.01$ & $0.01$ \\
FM    & QMC-FKS & $-2.63$ & $-3.05$ & $-2.00$ & $-1.91$ & $-2.41$  & $-3.37$  & $-2.42$ & $-2.73$ & $0.24$ & $-1.16$ & $0.33$ & $0.32$ & $0.01$ & $0.00$ \\
\addlinespace[2pt]
CMA-ES & N/A     & $-5.67$ & $-5.91$ & $-6.25$ & $-6.70$ & $-13.16$ & $-10.54$ & $-2.65$ & $-5.04$ & $-0.31$ & $-1.35$ & \multicolumn{1}{c}{--} & \multicolumn{1}{c}{--} & \multicolumn{1}{c}{--} & \multicolumn{1}{c}{--} \\
\addlinespace[2pt]
FM    & TCN (Ours) & \textcolor{teal}{$\mathbf{0.57}$} & \textcolor{teal}{$\mathbf{0.46}$} & \textcolor{teal}{$\mathbf{-0.42}$} & \textcolor{teal}{$\mathbf{-0.46}$} & \textcolor{teal}{$\mathbf{-1.03}$} & \textcolor{teal}{$\mathbf{-2.64}$} & \textcolor{teal}{$\mathbf{-0.90}$} & \textcolor{teal}{$\mathbf{-1.50}$} & \textcolor{teal}{$\mathbf{1.21}$} & \textcolor{teal}{$\mathbf{0.69}$} & \textcolor{teal}{$\mathbf{0.97}$} & \textcolor{teal}{$\mathbf{0.80}$} & $0.01$ & $0.07$ \\
\bottomrule
\end{tabular*}
\end{table*}

To support informative policy adaptation, the detector must recognize when the deployed system departs from the regime the policy was pretrained for. This section addresses \textbf{Q1}. Detectors are trained exclusively on Genesis trajectory pairs and evaluation uses MuJoCo for both reference and target windows. The evaluation physical parameters remain inside the broad prior but in its tails. For fairness, the alarm threshold is fixed at the \num{99}th percentile of anomaly logits on \num{2000} held-out nominal Genesis pairs, targeting \SI{99}{\percent} specificity in Genesis. Each robot--terrain setting contains \num{10000} nominal and \num{10000} anomalous MuJoCo target windows, together with \num{10000} independent nominal reference windows.

Table~\ref{tab:anomaly-detect} reports the true positive rate (TPR) and false positive rate (FPR) using three seeds. The measured FPR stays at \num{0.01} on \emph{Flat} and rises to at most \num{0.07} on \emph{Rough}, so the methods separate mainly on TPR: the TCN detector misses a few percent of MuJoCo test windows on \emph{Flat} and one in five on \emph{Rough}, MLP features miss more, and kernel features miss the most (\SIrange{30}{68}{\percent}). The detector schedules adaptation rather than producing improvement itself: a missed detection delays a useful update, whereas a false alarm spends a fine-tuning round and can cause policy drift.

We further measure how often adaptation is triggered in deployment. Over \num{9600} evaluation episodes per robot and environment, collected across the saved policy snapshots, the detector flags \SI{99.15}{\percent} / \SI{96.02}{\percent} of anomalous episodes on \emph{Flat} / \emph{Rough} for \texttt{Ant}, \SI{100.0}{\percent} / \SI{100.0}{\percent} for \texttt{Quadruped}, \SI{100.0}{\percent} / \SI{96.27}{\percent} for \texttt{Humanoid}, \SI{100.0}{\percent} / \SI{100.0}{\percent} for \texttt{Quadcopter}, and \SI{99.95}{\percent} / \SI{99.65}{\percent} for \texttt{Jackal}. Almost every anomalous episode therefore receives online fine-tuning, while false alarms remain infrequent (window-level FPR of \num{0.01} on \emph{Flat} and at most \num{0.07} on \emph{Rough}; Table~\ref{tab:anomaly-detect}), which limits how often a nominal window triggers unnecessary fine-tuning. The remaining misses concentrate on \emph{Rough}, where terrain variation acts as a nuisance.

\subsection{Neural Fidelity Calibration Results}

Addressing \textbf{Q2}, we evaluate how accurately NFC infers high-dimensional simulator parameters and residual fidelity, relative to the SBI baselines. The metric is the average log-density of the ground-truth parameters under each model's predicted posterior, $\log \boldsymbol{q}(\boldsymbol{\psi}^{*} \vert \boldsymbol{\tau})$, following~\citet{ramos2019bayessim}; Table~\ref{tab:anomaly-detect} reports it, and Fig.~\ref{fig:sim-posterior} visualizes two representative parameters per robot.

First, for simulator calibration, NFC attains the highest log-density on every robot and environment. Two trends locate the difficulty: for most methods accuracy drops from \emph{Flat} to \emph{Rough}, where perception enters the conditioning, and from \texttt{Quadcopter} to \texttt{Humanoid}, where $\vert\boldsymbol{\psi}\vert$ grows from \num{9} to \num{39}. The trend is not uniform---MDN-RFF on \texttt{Ant}, FM-FKS on \texttt{Quadruped}, and CMA-ES on \texttt{Humanoid} improve from \emph{Flat} to \emph{Rough}---but NFC retains the highest log-density in all ten settings. Among baselines, FM-MLP is the strongest in eight of the ten settings. With the encoder held fixed, FM-TCN exceeds MDN-TCN in all ten settings, which isolates the gain from flow matching; replacing the MLP by the TCN encoder adds the remaining margin, and kernel features (RFF, FKS) trail because a fixed random basis cannot adapt to high-dimensional conditioning.

Second, for residual fidelity, Fig.~\ref{fig:sim-posterior} (right) shows NFC's posteriors over residual position and residual height. Baselines are not shown on this target. Their accuracy already trails on the physical parameters (Table~\ref{tab:anomaly-detect}), and each baseline was evaluated on the targets stated in Sec.~\ref{sec:sim-settings}: kernel features approximate the conditioning with a fixed random basis that does not adapt to the target dimension, and CMA-ES maintains a covariance over the search space, which is $\mathcal{O}(d^2)$ in that dimension and requires multiple simulator rollouts per generation.

Third, latent NFC trained without the reconstruction objective performs on par with FM-MLP and well below the full model, confirming that reconstruction preserves the privileged information the latent needs---an effect also reported for teacher--student sim-to-real transfer~\citep{he2024bridging}. Two further observations qualify the comparison. CMA-ES is a strong optimizer in principle: in offline runs it does converge once the budget grows from about \num{10} to \num{100} simulation iterations, and on low-dimensional targets such as mass and inertia~\citep{sobanbabu2025sampling} it converges quickly. But every iteration re-runs the simulator, so \num{100} iterations per adaptation round are incompatible with online deployment, and on low-dimensional targets the neural estimators converge just as fast. Likewise, enlarging the MLP far beyond the BayesSim configuration yields modest gains. Our target is not inference accuracy alone but accurate posteriors at real-time cost, and on that trade-off NFC is the only method that meets both requirements.

\subsection{Policy Improvement Results}\label{sec:policy-improve-sim}

This section addresses \textbf{Q3}, evaluating policy fine-tuning performance with our NFC framework.
Building on the anomaly detection and fidelity calibration results, we assess how NFC enhances sim-to-sim adaptation. 

\textbf{Policy Baselines and Ablations.}
We benchmark:
\begin{enumerate}[label=(\arabic*), leftmargin=*]
    \item \sbf{PPO w/ NFC}: Our full method with both simulator parameter calibration and residual fidelity.
    \item \sbf{PPO w/o $\boldsymbol{\Delta}$Fidelity}: An ablation that disables the residual ($\boldsymbol{\Delta}$) fidelity module but retains simulator parameter inference.
    \item \sbf{PPO w/o NFC}: An ablation that disables NFC entirely, fine-tuning PPO directly on online execution data.
    \item \sbf{PPO w/ MDN-RFF}: A strong SBI method~\citep{ramos2019bayessim} using mixture density networks with RFF. We adjust it to consider the residual fidelity by inferring $ \Delta\boldsymbol{s} $ only and reusing the perceived map.
    \item \sbf{PPO w/ CMA-ES}: A non-neural optimization baseline that achieved strong results in mass--inertia estimation for quadrupeds~\citep{sobanbabu2025sampling}. It minimizes sim-to-real trajectory discrepancy by optimizing over the simulator parameters.
\end{enumerate}
The last two baselines are representative of neural and optimization-based approaches, respectively; others (e.g., MDN with our TCN encoder) performed consistently worse in preliminary evaluations and are omitted. PPO w/ NFC, PPO w/o $\Delta$Fidelity, and PPO w/ MDN-RFF share the hallucination policy; PPO w/o NFC and PPO w/ CMA-ES have no sampled domain to gate. To broaden the comparison, we also include a state-of-the-art MBRL algorithm:
\begin{enumerate}[label=(\arabic*), leftmargin=*, resume]
    \item \sbf{TD-MPC2}: A well-established MBRL method, which is pretrained in simulation and directly fine-tunes the dynamics model and policy solely from real-world data~\citep{hansen2024tdmpc}. We choose TD-MPC2 due to its data efficiency compared to model-free RL algorithms (e.g., PPO, SAC~\citep{haarnoja2018soft}).
\end{enumerate}

\begin{table*}[t]
\small
\centering
\caption{Normalized policy returns in the sim-to-sim evaluation. The comparison between PPO w/ NFC and PPO w/o $\Delta$Fidelity measures the combined contribution of residual fidelity beyond physical calibration. \textbf{\textcolor{teal}{Green}} and \textbf{Bold} indicate the best and second-best results.}
\label{tab:policy-improve}
\begin{tabular}{llcccccc}
\toprule
\multicolumn{2}{l}{Average Return $[0,1]$} & \begin{tabular}[c]{@{}c@{}}PPO\\ w/ NFC\end{tabular} & \begin{tabular}[c]{@{}c@{}}PPO\\ w/o $\Delta$Fidelity\end{tabular} & \begin{tabular}[c]{@{}c@{}}PPO\\ w/o NFC\end{tabular} & \begin{tabular}[c]{@{}c@{}}PPO\\ w/ MDN-RFF\end{tabular} & \begin{tabular}[c]{@{}c@{}}PPO\\ w/ CMA-ES\end{tabular} & TD-MPC2 \\
\midrule
\multirow{2}{*}{\texttt{Ant}} 
  & \emph{Flat}  & \textcolor{teal}{$\mathbf{0.98}$} & $\mathbf{0.88}$ & $0.64$ & $0.76$ & $0.72$ & $0.87$ \\
  & \emph{Rough} & \textcolor{teal}{$\mathbf{0.95}$} & $\mathbf{0.81}$ & $0.57$ & $0.62$ & $0.56$ & $0.75$ \\
\addlinespace[3pt]
\multirow{2}{*}{\texttt{Quadruped}} 
  & \emph{Flat}  & \textcolor{teal}{$\mathbf{0.67}$} & $0.33$ & $0.17$ & $0.21$ & $0.21$ & $\mathbf{0.42}$ \\
  & \emph{Rough} & \textcolor{teal}{$\mathbf{0.55}$} & $\mathbf{0.27}$ & $0.18$ & $0.20$ & $0.17$ & $0.18$ \\
\addlinespace[3pt]
\multirow{2}{*}{\texttt{Humanoid}} 
  & \emph{Flat}  & \textcolor{teal}{$\mathbf{0.96}$} & $\mathbf{0.89}$ & $0.55$ & $0.70$ & $0.67$ & $0.88$ \\
  & \emph{Rough} & \textcolor{teal}{$\mathbf{0.93}$} & $\mathbf{0.90}$ & $0.58$ & $0.61$ & $0.58$ & $0.88$ \\
\addlinespace[3pt]
\multirow{2}{*}{\texttt{Quadcopter}} 
  & \emph{Flat}  & \textcolor{teal}{$\mathbf{0.82}$} & $\mathbf{0.75}$ & $0.67$ & $0.71$ & $0.69$ & $0.72$ \\
  & \emph{Rough} & \textcolor{teal}{$\mathbf{0.81}$} & $\mathbf{0.69}$ & $0.33$ & $0.68$ & $0.66$ & $0.67$ \\
\addlinespace[3pt]
\multirow{2}{*}{\texttt{Jackal}} 
  & \emph{Flat}  & \textcolor{teal}{$\mathbf{0.87}$} & $\mathbf{0.67}$ & $0.23$ & $0.53$ & $0.52$ & $0.41$ \\
  & \emph{Rough} & \textcolor{teal}{$\mathbf{0.82}$} & $\mathbf{0.69}$ & $0.20$ & $0.52$ & $0.50$ & $0.41$ \\
\bottomrule
\end{tabular}
\end{table*}

\textbf{Results.}
Table~\ref{tab:policy-improve} presents the normalized average return across five robot embodiments. The average return, computed as the cumulative reward, is normalized to $ [0,1] $ using the running bounds. Our approach, PPO w/ NFC, consistently outperforms all baselines and ablations. Removing the residual fidelity module (PPO w/o $\Delta$Fidelity) lowers return in all ten settings; the drop is largest on \texttt{Quadruped} (\num{0.34} on \emph{Flat}, \num{0.28} on \emph{Rough}) and smallest on \texttt{Humanoid} (\num{0.07} and \num{0.03}), so the contribution of residual fidelity varies by embodiment rather than growing with robot complexity. This highlights the importance of capturing unmodeled discrepancies in environment and dynamics. Completely disabling NFC (PPO w/o NFC) causes the most significant performance loss, reinforcing that NFC narrows the domain randomization range to match real-world distributions more precisely. MDN-RFF and CMA-ES yield smaller and less consistent gains over PPO w/o NFC than NFC does, consistent with their lower inference accuracy.

TD-MPC2 demonstrates strong performance as expected from an MBRL method. It outperforms PPO w/o NFC in nine of ten settings (tying on \texttt{Quadruped}, \emph{Rough}) and PPO w/o $\Delta$Fidelity in one (\texttt{Quadruped}, \emph{Flat}: \num{0.42} vs.\ \num{0.33}), but never PPO w/ NFC, underscoring the contribution of residual fidelity modeling. This comparison also reveals a critical distinction: without NFC, TD-MPC2 fine-tunes entirely in the real world, whereas our framework can fine-tune in the simulation. Moreover, its learned perception-aware dynamics model runs at only \SI{6}{\hertz} versus \SI{25}{\hertz} for PPO-based methods, limiting real-time use.

\section{Real-World Experiment}\label{sec:real-world}
Revisiting \textbf{Q2} and \textbf{Q3}, we validate our framework on the Clearpath Jackal, a differential-drive robot, under challenging and anomalous real-world conditions. Building upon our simulation experiments, we evaluate sim-to-real transfer across laboratory settings (\emph{Flat}) and complex natural terrains (\emph{Rough}). This section highlights the challenges associated with real-world deployment.

\subsection{Sim-to-Real Experimental Settings}
This section details the deployment setup, experimental design, and evaluation metrics, highlighting the practicality of our evaluation for real-world robotics.
 
\begin{figure}[t]
\vspace*{0.04in}
\centering
{\setlength{\fboxsep}{0pt}\setlength{\fboxrule}{0.4pt}\fbox{\includegraphics[width=\dimexpr\linewidth-2\fboxrule\relax]{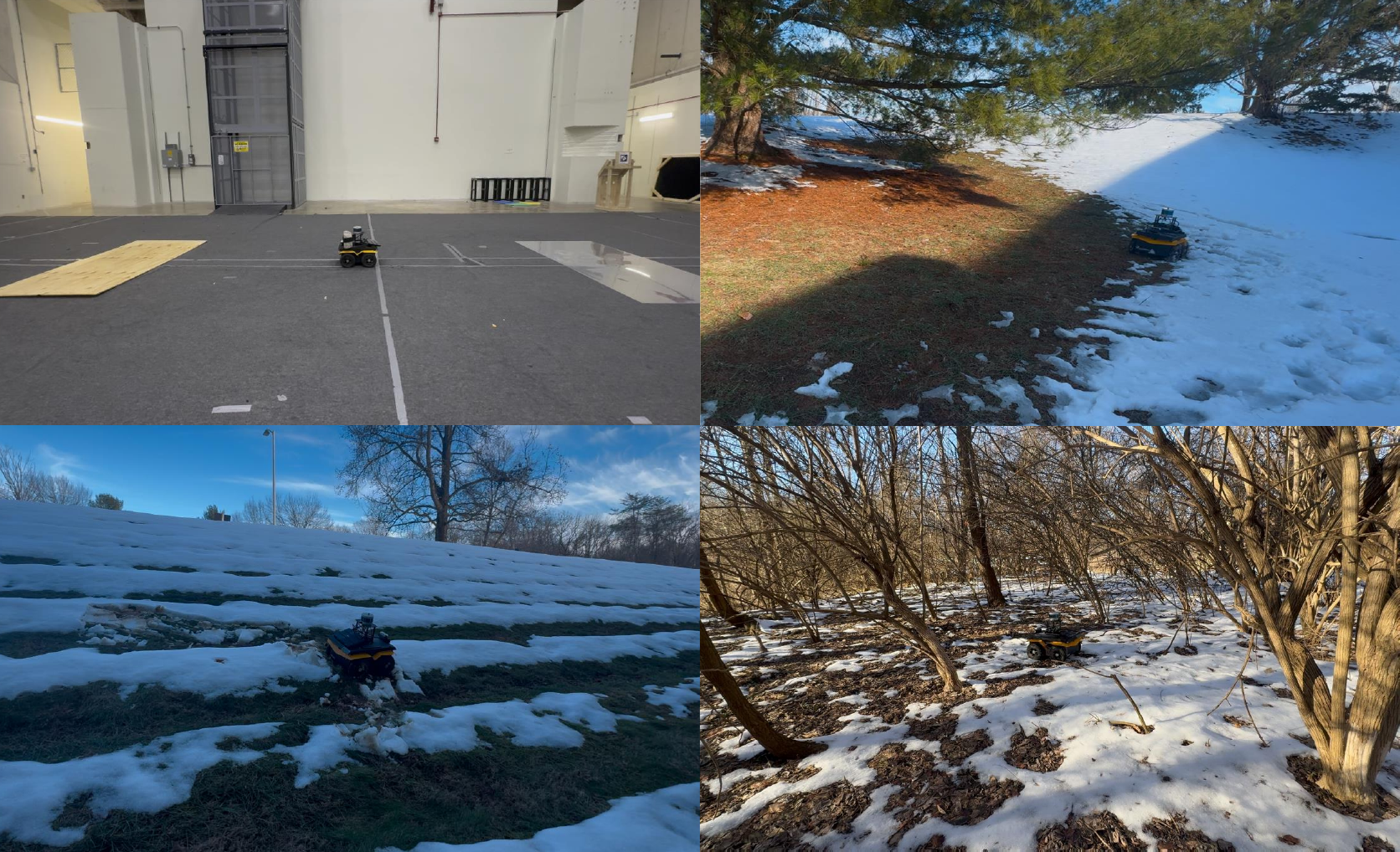}}}
\caption{Real-world evaluation scenarios for a Jackal with a broken front-left wheel axle. Laboratory conditions vary surface material and the presence of a loosely attached rock; outdoor conditions combine snow, wet grass, mud, and elevation changes. The rock is used only in the laboratory scenarios.}
\label{fig:real-world-envs}
\end{figure}

\textbf{Robot and Environment.}
Pretraining follows the \texttt{Jackal} configuration of Sec.~\ref{sec:sim-exp}. The Jackal robot is equipped with a RealSense D435i depth camera, a 16-beam Velodyne LiDAR, and a MicroStrain 3DM-GX5-25 IMU. Depth images (\SI{30}{\hertz}) and LiDAR-inertial odometry (\SI{200}{\hertz}) are fused on the onboard Jetson Orin to produce real-time elevation maps~\citep{RAL18-elemap,yu2023mimu}. Policy optimization is offloaded to a laptop with an RTX~3080~Ti~Mobile GPU. Fig.~\ref{fig:real-world-envs} shows the real-world testing scenarios, covering a range of laboratory surfaces (\emph{Flat}) and outdoor terrain with elevation changes and complex physical properties (\emph{Rough}). First, in the \emph{Flat} scenarios, we design four challenges, all with a broken front-left wheel axle:
\begin{enumerate}[label=(\arabic*), leftmargin=*]
    \item standard flat surface,
    \item flat surface with an added rock on the robot (not tightly attached),
    \item flat surface with different materials,
    \item flat surface with different materials with an added rock on the robot (not tightly attached).
\end{enumerate}
The rock is included only in \emph{Flat} environments, as its mass and inertia introduce significant challenges even for flat-surface navigation and would likely make the mission impossible to complete in \emph{Rough} environments. \emph{Rough} represents outdoor terrain, including snowy and muddy terrain, wet grass, and elevated surfaces. We evaluate the following challenges, all with a broken front-left wheel axle:
\begin{enumerate}[label=(\arabic*), leftmargin=*, resume]
    \item wet grass and snowy terrain,
    \item elevated terrain with wet grass and snow,
    \item muddy and snowy terrain in a forest.
\end{enumerate}
In each scenario, the robot is tasked with following a figure-\num{8} trajectory with a \SI{3}{\meter} diameter. Each controller is tested \num{9} times in each scenario.

\textbf{Baselines and Ablations.}
We evaluate both parameter inference and policy adaptation in the real world. Following a similar procedure as in the sim-to-sim experiments, we compare \sbf{NFC} ({FM} with {TCN}) against \sbf{FM-MLP}, \sbf{FM-RFF}, \sbf{FM-FKS}, \sbf{MDN-RFF}, \sbf{MDN-TCN}, and \sbf{CMA-ES} for simulator calibration. For policy improvement, we evaluate \sbf{PPO w/ NFC} alongside its ablations, \sbf{PPO w/o NFC} and \sbf{PPO w/o $ \Delta $Fidelity}, as well as baselines \sbf{MDN-RFF} and \sbf{CMA-ES}, and the MBRL method \sbf{TD-MPC2}. TD-MPC2 is pretrained in simulation and directly fine-tunes the dynamics model solely from real-world robot trajectories. Additionally, we include a high-rate motion primitive planner, \sbf{Falco}~\citep{Zhang2020Falco}, with tuned PID control (\SI{200}{\hertz}) and access to elevation maps. It serves as a strong baseline for comparison in reactive navigation scenarios~\citep{cao2023granularity, Yang-RSS-23}.

\textbf{Evaluation Metrics.}
We follow previous works~\citep{siva2021enhancing, yu2024adaptive} and evaluate performance using success rate, trajectory ratio, orientation jerk $ \left| \frac{\partial^2 \omega}{\partial t^2} \right| $, and position jerk $ \left| \frac{\partial a}{\partial t} \right| $, where $ \omega $ and $ a $ represent angular velocity and linear acceleration, respectively. These motion-stability indicators capture sudden pose changes, which matter for safety. The trajectory ratio is the robot's traveled path length divided by the target path length over successfully completed laps; \num{1.0} is ideal, and larger values indicate detours or oscillations (lower is better).
A lap counts as successful when the robot reaches every waypoint within \SI{0.25}{\meter} and completes the course without rollover or human intervention.

Across the \num{2160} anomalous trajectories of the real-world runs, the detector flagged every one, so adaptation was triggered in all deployments. The following sections therefore report posterior inference and policy improvement without a separate detection comparison.

\subsection{Neural Fidelity Calibration Results}

\begin{figure*}[t!]
\centering
{\setlength{\fboxsep}{0pt}\setlength{\fboxrule}{0.4pt}\fbox{\includegraphics[width=0.90\linewidth]{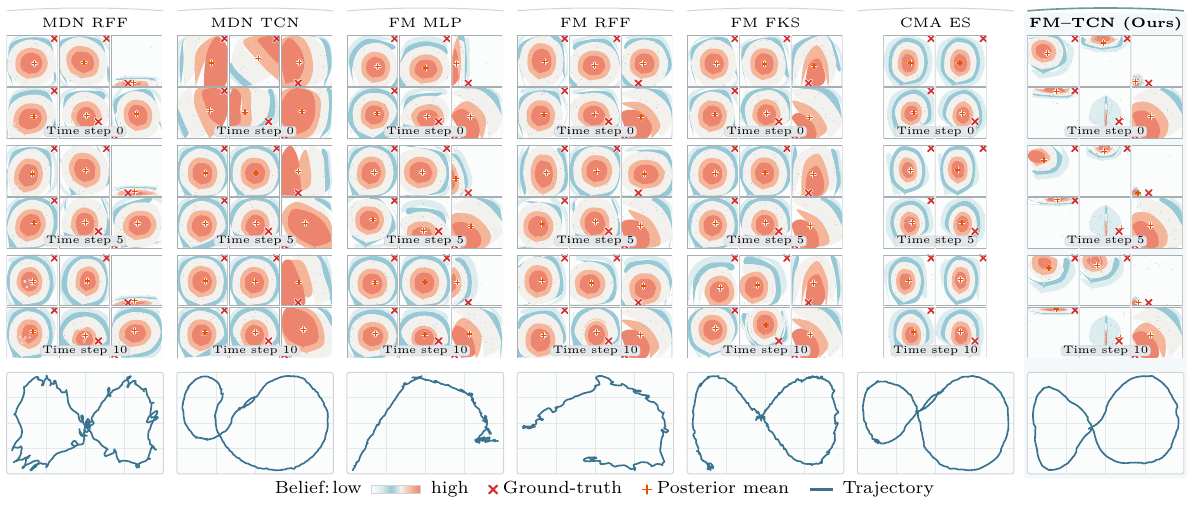}}}
\caption{Neural Fidelity Calibration results for a wheeled robot navigating real-world snowy terrain with a broken front-left wheel axle. Demonstrated trajectories are visualized alongside inferred parameters. For each time step (one history window), each method block shows a $2\times3$ panel grid: the first two columns are joint posteriors over wheel friction ($x$-axis) and damping ($y$-axis) for the four wheels (top row: rear-right, rear-left; bottom row: front-right, front-left), and the third column shows the inferred residual dynamics ($\Delta x, \Delta y$; top) and residual terrain height ($\Delta z$; bottom); CMA-ES has no residual column. Real-world parameters cannot be measured directly, so the red crosses labeled ``Ground-truth'' are estimated references.}
\label{fig:real-world-posterior}
\end{figure*}

We present the comparison of simulator calibration and the results of our residual fidelity in Fig.~\ref{fig:real-world-posterior}. The calibrated damping of the front-left wheel averages \num{0}, consistent with an undriven, free-spinning wheel, while the friction across all four wheels averages \num{0.4}, reflecting the snowy terrain. Competing methods, despite being trained on the same dataset, produce diffuse posteriors that do not single out the front-left wheel. This result aligns with our sim-to-sim experiments, reinforcing our method's superiority in high-dimensional parameter spaces. We also present the results of the residual fidelity module, with residual dynamics shown in Fig.~\ref{fig:real-world-posterior} and residual environment in Fig.~\ref{fig:offroad-res-env}. Measured ground truth for the residual fidelity is unavailable in the real world (the red crosses in Fig.~\ref{fig:real-world-posterior} are estimates), so its effect is measured through policy improvement in the following section.

\subsection{Policy Improvement Results}

\begin{figure*}[t]
\centering
{\setlength{\fboxsep}{0pt}\setlength{\fboxrule}{0.4pt}\fbox{\includegraphics[width=0.97\linewidth]{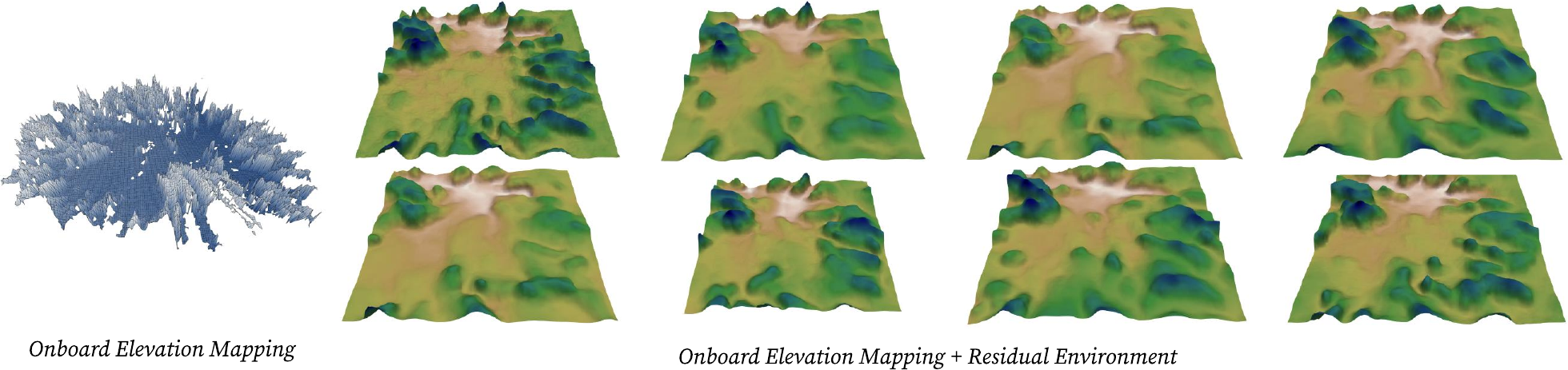}}}
\caption{Residual environment inference for real-world adaptation. NFC conditions on robot execution and the perceived elevation map $\hat{\boldsymbol{e}}$ to sample corrections $\Delta\boldsymbol{e}_n$. The eight reconstructions $\boldsymbol{e}_n = \hat{\boldsymbol{e}} + \Delta\boldsymbol{e}_n$ are demonstrative simulation domains for policy fine-tuning.}
\label{fig:offroad-res-env}
\end{figure*}

\begin{table*}[t]
\small
\centering
\caption{Real-world policy improvement results for our method (PPO w/ NFC), its ablations, and baselines. A Clearpath Jackal with a broken front-left wheel axle is evaluated across four \emph{Flat} (lab surface) and three \emph{Rough} (outdoor terrain) scenarios, demonstrating the importance of residual fidelity on policy adaptation under anomalous conditions. \textbf{\textcolor{teal}{Green}} and \textbf{Bold} indicate the best and second-best results.}
\label{tab:real-world-policy-improve}
\begin{tabular}{llccccccc}
\toprule
\multicolumn{2}{l}{} & \begin{tabular}[c]{@{}c@{}}PPO\\ w/ NFC\end{tabular} & \begin{tabular}[c]{@{}c@{}}PPO\\ w/o $\Delta$Fidelity\end{tabular} & \begin{tabular}[c]{@{}c@{}}PPO\\ w/o NFC\end{tabular} & \begin{tabular}[c]{@{}c@{}}PPO\\ w/ MDN-RFF\end{tabular} & \begin{tabular}[c]{@{}c@{}}PPO\\ w/ CMA-ES\end{tabular} & TD-MPC2 & Falco \\
\midrule
\multirow{2}{*}{\begin{tabular}[c]{@{}l@{}}Success Rate\\ (\%) ($\uparrow$)\end{tabular}} 
  & \emph{Flat}  & \textcolor{teal}{$\mathbf{100}$} & \textcolor{teal}{$\mathbf{100}$} & \textcolor{teal}{$\mathbf{100}$} & \textcolor{teal}{$\mathbf{100}$} & \textcolor{teal}{$\mathbf{100}$} & \textcolor{teal}{$\mathbf{100}$} & \textcolor{teal}{$\mathbf{100}$} \\
  & \emph{Rough} & \textcolor{teal}{$\mathbf{70}$}  & $\mathbf{52}$  & $26$  & $48$  & $44$  & $41$  & $30$ \\
\addlinespace[3pt]
\multirow{2}{*}{\begin{tabular}[c]{@{}l@{}}Traj. Ratio ($\downarrow$)\end{tabular}}
  & \emph{Flat}  & $1.80$                            & $1.80$                           & $1.90$                           & $\mathbf{1.53}$                  & $1.99$                           & \textcolor{teal}{$\mathbf{1.50}$} & $1.60$ \\
  & \emph{Rough} & \textcolor{teal}{$\mathbf{2.10}$} & $3.20$                           & $3.70$                           & $2.40$                           & $\mathbf{2.35}$                  & $2.70$                            & $3.60$ \\
\addlinespace[3pt]
\multirow{2}{*}{\begin{tabular}[c]{@{}l@{}}Orien. Jerk ($\downarrow$)\\ (\si{\radian\per\second\cubed})\end{tabular}} 
  & \emph{Flat}  & \textcolor{teal}{$\mathbf{2.98}$} & $\mathbf{3.50}$                  & $14.18$                          & $6.14$                           & $7.96$                           & $6.95$                            & $41.90$ \\
  & \emph{Rough} & \textcolor{teal}{$\mathbf{3.19}$} & $\mathbf{8.49}$                  & $26.99$                          & $9.81$                           & $10.38$                          & $8.98$                            & $41.93$ \\
\addlinespace[3pt]
\multirow{2}{*}{\begin{tabular}[c]{@{}l@{}}Pos. Jerk ($\downarrow$)\\ (\si{\meter\per\second\cubed})\end{tabular}} 
  & \emph{Flat}  & \textcolor{teal}{$\mathbf{4.36}$} & $9.32$                           & $8.41$                           & $6.92$                           & $7.79$                           & $\mathbf{5.27}$                   & $6.95$ \\
  & \emph{Rough} & \textcolor{teal}{$\mathbf{6.26}$} & $12.46$                          & $12.42$                          & $9.39$                           & $11.44$                          & $18.46$                           & $\mathbf{8.87}$ \\
\bottomrule
\end{tabular}
\end{table*}

As shown in Table~\ref{tab:real-world-policy-improve}, the real-world scenarios present significant challenges for wheeled robots. We organize the analysis into three parts, highlighting (i) the impact of our Neural Fidelity Calibration, (ii) the contribution of the Residual Fidelity module, and (iii) the performance of baseline methods.

First, NFC plays a critical role in enabling effective online policy adaptation. When NFC is ablated and PPO is fine-tuned solely using real-world data (PPO w/o NFC), performance gains are minimal across all metrics. In the \emph{Flat} setting, PPO achieves a \SI{100}{\percent} success rate but exhibits frequent trajectory oscillations and high jerk, indicating that the policy handles the anomaly in outcome but not in motion quality. In \emph{Rough} scenarios, the absence of simulator calibration leads to degraded behavior---most notably, repetitive back-and-forth movements and harsh jerks---suggesting the policy gets stuck in local minima. One noted limitation of NFC appears in the \emph{Flat} setting, where the trajectory ratio remains suboptimal despite successful adaptation, due to the absence of trajectory length penalties in the reward function~\citep{yu2024adaptive}. 

Second, we assess the contribution of $\Delta$Fidelity. Fig.~\ref{fig:offroad-res-env} illustrates the onboard perceived environment, which is noisy and incomplete due to complex geometry and material properties outdoors. One potential alternative is to directly inject perceived maps into simulation using linear interpolation. However, this strategy fails to recover sufficient structure, as evidenced by the performance of PPO w/o $\Delta$Fidelity, which improves over PPO w/o NFC (\SI{52}{\percent} vs.\ \SI{26}{\percent} \emph{Rough} success) but stays well below the full method. On one hand, PPO w/o $\Delta$Fidelity has far lower orientation jerk than PPO w/o NFC (though similar position jerk), as it still benefits from simulator calibration fine-tuning. On the other hand, its remaining gap to PPO w/ NFC is attributable to the residual dynamics and perception uncertainty that only $\Delta$Fidelity models.

Third, we compare to strong baselines. Falco attains low position jerk from its \SI{200}{\hertz} control loop but the highest orientation jerk, as its primitives do not account for the asymmetric traction of the broken axle. The MBRL method TD-MPC2, which is fine-tuned using only online execution data, also demonstrates advantages over PPO w/o NFC. As observed in the sim-to-sim experiments, TD-MPC2 benefits from its sample efficiency but suffers from computational inefficiencies, especially when incorporating perception into state propagation. The results further highlight the importance of our NFC module. Initially, PPO alone does not outperform TD-MPC2, but after fine-tuning with NFC, PPO substantially surpasses TD-MPC2, demonstrating the effectiveness of NFC-driven policy adaptation in real-world environments.

In summary, NFC raises \emph{Rough} success from \SI{26}{\percent} (PPO w/o NFC) to \SI{70}{\percent} while reducing orientation jerk by \SI{88}{\percent}, handling a broken axle and unstructured terrain together.

\section{Limitations and Future Work}
\label{sec:limitations}
Three limitations bound the current framework.
First, detection rests on the contrast between a reference and a target window, so a change that develops slowly relative to the window length produces little signal and is absorbed only in the post-mission offline phase. Second, physical parameters and residuals need not be uniquely identifiable from a short trajectory, so inferred values are best read together with the downstream policy result rather than as measured physical properties. Third, each adaptation round updates the policy from its current parameters, so returning to an earlier regime re-estimates the domain distribution without by itself restoring the earlier behavior. Maintaining regime-indexed policies selected by the NFC posterior is a plausible extension.

Our approach also assumes a black-box simulator. Estimating the gradient of robot trajectories with respect to the parameters through perturbation theory, or using differentiable simulators~\citep{taichi, Lv-RSS-23}, could accelerate inference, although exploding gradients~\citep{warp2022} and partial observability remain open issues. Multi-modal conditioning, such as image segmentation, is another direction: decomposing the conditioning across perception modules would add flexibility, but it assumes independence between them, which need not hold in practice.

Finally, many deployed systems pair a performance policy with a dedicated recovery policy that intervenes when risk measures indicate imminent failure. Our framework is compatible with this paradigm, since the anomaly detector could trigger recovery as readily as adaptation, but we kept the two apart so that the reported gains are attributable to policy fine-tuning. Integrating them, with each recovery activation triggering an NFC update of the performance policy, is a natural next step.

\section{Conclusion}
\label{sec:conclusion}
We have presented Neural Fidelity Calibration (NFC) to simultaneously calibrate simulator physics and residual fidelity parameters, facilitating policy fine-tuning online. Our novel residual fidelity consists of residual dynamics and residual environment, where the former captures the simulation shift relative to real-world dynamics and the latter quantifies uncertainty in environmental perception. Leveraging NFC's superior posterior inference, we enhance real-world adaptability by integrating sequential latent NFC and optimistic exploration. Extensive sim-to-sim and sim-to-real experiments validate NFC's effectiveness: on real rough terrain with a broken wheel axle, residual fidelity raises success from \SI{52}{\percent} to \SI{70}{\percent}, and the full framework outperforms both classic and learning-based approaches.

\begin{dci}
The authors declared no potential conflicts of interest with respect to the research, authorship, and/or publication of this article.
\end{dci}

\bibliographystyle{ijrr/SageH}

\appendix

\section{Kernel Features}\label{apdx:kernel-feats}
Kernel methods have achieved success in feature extraction for simulation-based inference~\citep{ramos2019bayessim}. Here, we briefly review the radial kernel and introduce the more powerful fast kernel slicing method.

\subsection{Radial Kernel}
We introduce the basic ideas of using radial kernels for feature representation. 
For inputs $ x \in \mathbb{R}^d $ and $ y \in \mathbb{R}^d $, the radial basis function (RBF) family uses radial kernels $ K(x,y) = F(\|x - y\|) $ with a basis function $ F(\cdot) : \mathbb{R}_{\geq 0} \rightarrow \mathbb{R} $. We list some well-known kernels:
\begin{enumerate}[leftmargin=*, label=(\arabic*)]
    \item Gaussian: $ F(x) = \exp\left( - \frac{x^2}{2 \sigma^2} \right) $.
    \item Laplacian: $ F(x) = \exp\left( - \alpha x \right) $.
    \item Matérn: $ F(x) = \frac{2^{1-\nu}}{\Gamma(\nu)} \left( \frac{\sqrt{2\nu}}{\beta} x \right)^{\nu} K_\nu \left( \frac{\sqrt{2\nu}}{\beta} x \right) $.
\end{enumerate}

For scaling to large numbers of training samples or high-dimensional states, Random Fourier Feature (RFF) and Fast Kernel Slicing (FKS) can approximate RBF kernels.

\subsubsection{Random Fourier Feature}

Let $ K : \mathbb{R}^d \times \mathbb{R}^d \to \mathbb{R} $ be given by $ K(x, y) = \Psi(x - y) $, a bounded shift-invariant positive definite kernel with $ K(x, x) = \Psi(0) = 1 $. Then it holds by Bochner's theorem that there exists a probability measure $ \mu $ on $ \mathbb{R}^d $ such that
\begin{equation}
\Psi(x - y) = \mathbb{E}_{\substack{v \sim \mu \\ b \sim U[0, 2\pi]}} \Big[ 2 \cos(\langle x, v \rangle + b) \cos(\langle y, v \rangle + b) \Big].
\end{equation}

Random Fourier Feature simply puts $ y = 0 $,
\begin{equation}
\begin{aligned}
\Psi(x) = \mathbb{E}_{v \sim \mu} \left[ \exp\left( i \langle x, v \rangle \right) \right].
\end{aligned}
\end{equation}

Using Euler's formula and taking the real part, noting that $\Psi$ is real-valued, we obtain
\begin{equation}
\Psi(x) = \mathbb{E}_{v \sim \mu} \left[ \cos\left( \langle x, v \rangle \right) \right].
\end{equation}

To subsample the expectations, let $ \{v_1, \dots, v_D\} $ be QMC points from $ \mu $ and $ \{b_1, \dots, b_D\} $ be i.i.d. samples from $ U[0, 2\pi] $. We then have
\begin{equation}
\mathtt{RFF}(x) = \sqrt{\tfrac{1}{D}} \left[ \cos(\langle x, v_p \rangle + b_p),\ \sin(\langle x, v_p \rangle + b_p) \right]_{p=1}^{D} .
\end{equation}

\subsection{Fast Kernel Slicing}\label{apdx:fast-fourier-sum}
We briefly delineate fast kernel slicing methods~\citep{johannes2024slice}, which reduce high-dimensional problems to the one-dimensional case by projecting the points of interest onto lines in directions $ \xi $ that are uniformly distributed on the unit sphere $ \mathbb{S}^{d-1} \subset \mathbb{R}^{d} $. In this paper, we assume the kernel $ K : \mathbb{R}^{d} \times \mathbb{R}^{d} \rightarrow \mathbb{R} $ has the form
\begin{equation}
    K(x, y) = \mathbb{E}_{\xi \sim \mathcal{U}_{\mathbb{S}^{d-1}}} \left[ k(\langle \xi, x \rangle, \langle \xi, y \rangle) \right],
\end{equation}
with a one-dimensional kernel $ k : \mathbb{R} \times \mathbb{R} \rightarrow \mathbb{R} $ and the uniform distribution $ \mathcal{U}_{\mathbb{S}^{d-1}} $ on the sphere. For a radial kernel $ K(x, y) = F(\|x - y\|) $ with basis function $ F: \mathbb{R}_{\geq 0} \rightarrow \mathbb{R} $, the one-dimensional kernel is also radial, $ k(a, b) = f(|a - b|) $, with a one-dimensional basis function $ f $. Define $ L_{\mathrm{loc}}^{\infty} := \{ f : \mathbb{R}_{\geq 0} \rightarrow \mathbb{R} \,:\, f\vert_{[0,s]} \in L^{\infty}([0,s]) \text{ for all } s > 0 \} $. Then $ F $ and $ f $ are related by the \emph{generalized Riemann--Liouville fractional integral} transform $ \mathcal{S}_d : L_{\mathrm{loc}}^{\infty} \rightarrow L_{\mathrm{loc}}^{\infty} $, $ \mathcal{S}_d(f) = F $, with
\begin{equation}\label{eqn:kernel-sum}
    F(s) = \frac{2\, \Gamma(\frac{d}{2})}{\sqrt{\pi}\, \Gamma(\frac{d-1}{2})} \int_{0}^{1} f(ts)\, (1 - t^2)^{\frac{d-3}{2}}\, \mathrm{d} t .
\end{equation}
Given $ F $, the one-dimensional basis function $ f $ can be computed explicitly if either
\begin{enumerate}[label=(\arabic*), leftmargin=*]
    \item $ F $ is analytic with a globally convergent Taylor series at $ 0 $, $ F(s) = \sum_{m=0}^{\infty} a_m s^m $; or
    \item $ F $ is continuous, bounded, and positive definite, i.e., for all $ n \in \mathbb{N} $, points $ x_1, \dots, x_n \in \mathbb{R}^{d} $, and coefficients $ c_1, \dots, c_n \in \mathbb{R} $, it holds that $ \sum_{i,j=1}^{n} c_i c_j F(\|x_i - x_j\|) \geq 0 $.
\end{enumerate}

Drawing $ D $ directions $ \xi_1, \dots, \xi_D $ from $ \mathcal{U}_{\mathbb{S}^{d-1}} $, we obtain $ D $ features from the projections and the one-dimensional basis function with weights $ w_p $,
\begin{equation}\label{eqn:fast-features}
    \mathtt{FKS}(x) = \{ w_p f(\langle \xi_p, x \rangle) \}_{p=1}^{D},
\end{equation}
whose computational complexity is $ \mathcal{O}(D d) $, where $ d $ is the dimension of the input $ x $. The expected slicing error is bounded by $ \mathcal{O}(1 / \sqrt{D}) $. For a fair comparison to RFF, we assign uniform weights $ w_p $.

\section{Policy Parameterization}\label{apdx:policy-param}
Hyperparameters for PPO and TD-MPC2 are summarized in Tables~\ref{tab:policy-hyperparameters} and~\ref{tab:tdmpc2-hyperparameters}. The hallucination policy is a separate MLP that receives the history embedding $ \mathcal{H}(\hat{\boldsymbol{\tau}}) $ and outputs $ D = \vert \boldsymbol{\psi} \vert + \vert \boldsymbol{\phi}_{z} \vert $ gates, one per coordinate of the joint NFC sample $ \boldsymbol{x} = [\boldsymbol{\psi}, \boldsymbol{\phi}_{z}] $, with $ \vert \boldsymbol{\phi}_{z} \vert $ equal to the encoder's output dimension.

\begin{table}[htbp]
\small
\centering
\caption{Hyperparameters for the PPO Policy.}
\label{tab:policy-hyperparameters}
\begin{tabular*}{\linewidth}{@{\extracolsep{\fill}} lr @{}}
\toprule
\textbf{Hyperparameter} & \textbf{Value} \\
\midrule
\multicolumn{2}{@{}l}{\textbf{PPO}} \\
\quad Discount Factor & $0.99$ \\
\quad Learning Epoch & $2$ \\
\quad Generalized Advantage Estimation & $0.95$ \\
\quad Entropy Regularization Coefficient & $0.02$ \\
\quad PPO Clip Parameter & $0.20$ \\
\quad Optimizer & AdamW \\
\quad Learning Rate & $\num{5e-4}$ \\
\quad Minibatch Size & $\num{32768}$ \\
\quad Mini Epochs & $4$ \\
\midrule
\multicolumn{2}{@{}l}{\textbf{Hallucination Policy}} \\
\quad MLP Architecture & $[256, 256, 256]$ \\
\quad Activation & Tanh \\
\midrule
\multicolumn{2}{@{}l}{\texttt{Ant}} \\
\quad MLP Architecture & $[512, 256, 128]$ \\
\quad Activation & ELU \\
\midrule
\multicolumn{2}{@{}l}{\texttt{Quadruped}} \\
\quad Entropy Regularization Coefficient & $0.001$ \\
\quad MLP Architecture & $[512, 256, 128]$ \\
\quad Activation & ELU \\
\midrule
\multicolumn{2}{@{}l}{\texttt{Humanoid}} \\
\quad Entropy Regularization Coefficient & $0.008$ \\
\quad MLP Architecture & $[800, 400, 200]$ \\
\quad Activation & ELU \\
\midrule
\multicolumn{2}{@{}l}{\texttt{Quadcopter}} \\
\quad Entropy Regularization Coefficient & $0.016$ \\
\quad MLP Architecture & $[512, 512, 256]$ \\
\quad Activation & ELU \\
\midrule
\multicolumn{2}{@{}l}{\texttt{Jackal}} \\
\quad Entropy Regularization Coefficient & $0.005$ \\
\quad MLP Architecture & $[1024, 512, 256]$ \\
\quad Activation & Tanh \\
\bottomrule
\end{tabular*}
\end{table}

\begin{table}[htbp]
\small
\centering
\caption{Hyperparameters for the TD-MPC2 Policy.}
\label{tab:tdmpc2-hyperparameters}
\begin{tabular*}{\linewidth}{@{\extracolsep{\fill}} lr @{}}
\toprule
\textbf{Hyperparameter} & \textbf{Value} \\
\midrule
\multicolumn{2}{@{}l}{\textbf{TD-MPC2}} \\
\quad Encoder MLP & $[256, 256] \to \mathbb{R}^{512}$ \\
\quad Encoder Activation & Mish, SimNorm \\
\quad Dynamics MLP & $[512, 512]$ \\
\quad Dynamics Activation & Mish, SimNorm \\
\quad Reward MLP & $[512, 512] \to \mathbb{R}^{101}$ \\
\quad Reward Activation & Mish \\
\quad Policy MLP & $[512, 512]$ \\
\quad Policy Activation & Mish \\
\quad Q Ensemble MLP & $[512, 512] \to \mathbb{R}^{101}$ \\
\quad Q Ensemble Activation & Mish \\
\quad Q Ensemble Number & $5$ \\
\quad Minibatch Size & $\num{32768}$ \\
\quad Mini Epochs & $4$ \\
\quad Optimizer & AdamW \\
\quad Learning Rate & $\num{3e-4}$ \\
\quad Gradient Clip Norm & $20$ \\
\midrule
\multicolumn{2}{@{}l}{\textbf{MPPI}} \\
\quad Horizon & $6$ \\
\quad Iterations & $6$ \\
\quad Population Size & $512$ \\
\quad Number of Elites & $64$ \\
\quad Maximum std. & $1.0$ \\
\quad Temperature & $0.6$ \\
\bottomrule
\end{tabular*}
\end{table}

\subsection{Reward and Episode Settings}
All robots run at \SI{25}{\hertz} and receive actions clipped to $[-1,1]$. Let $\boldsymbol{v}$ and $\boldsymbol{\omega}$ denote body linear and angular velocity, $\boldsymbol{g}$ the projected gravity, $\boldsymbol{a}$ the action, and $d$ the distance to the goal.

\texttt{Quadruped} and \texttt{Humanoid} use a velocity-tracking reward,
\begin{equation}
\begin{aligned}
r = & \exp\!\big(-\|\boldsymbol{v}^{\mathrm{cmd}}_{xy} - \boldsymbol{v}_{xy}\|^2 / 0.25\big) \\
    & + \tfrac{1}{2}\exp\!\big(-(\omega^{\mathrm{cmd}}_{z} - \omega_{z})^2 / 0.25\big) - c_{\mathrm{reg}},
\end{aligned}
\end{equation}
where the regularizer $c_{\mathrm{reg}}$ penalizes vertical and roll--pitch motion, joint torques and accelerations, action rate, body tilt $\|\boldsymbol{g}_{xy}\|^2$, and undesired contacts on the trunk or pelvis.

\texttt{Jackal} is rewarded for progress toward the goal, $5(d_{t-1} - d_t)$, with a bonus while $d_t < \SI{0.2}{\meter}$, and penalties on body tilt and on abrupt changes in linear and angular velocity. \texttt{Quadcopter} is rewarded by $15\,(1 - \tanh(d/0.8))$ with penalties on linear and angular velocity. \texttt{Ant} is rewarded for progress toward a distant target, with an upright bonus, a heading term, and costs on action magnitude, energy, and joint-limit proximity.

\bibliography{references}

\end{document}